\documentclass[11pt, a4paper]{lumia}

\usepackage[sort&compress]{natbib}
\usepackage{xspace}

\theoremstyle{plain}

\newtheorem*{proposition*}{Proposition}

\theoremstyle{definition}

\theoremstyle{definition}

\def\eqref#1{equation~\ref{#1}}

\usepackage{graphicx}           
\usepackage{tikz}               
\usepackage{arydshln}           
\usepackage[edges]{forest}      

\usepackage{url}                
\usepackage{xurl}               

\usepackage{array}              
\usepackage{longtable}          
\usepackage{multirow}           
\usepackage{makecell}           
\usepackage{ragged2e}           

\usepackage{mathtools}          
\usepackage{nicefrac}           

\usepackage{algorithm}          
\usepackage{algorithmicx}       
\usepackage{algpseudocode}      
\usepackage{listings}           

\usepackage{subcaption}         
\usepackage{wrapfig}            
\usepackage[export]{adjustbox}  

\usepackage{changes}            
\usepackage{xspace}             
\usepackage[normalem]{ulem}     
\usepackage{CJKutf8}            

\usepackage[tikz]{bclogo}       
\usepackage[framemethod=tikz]{mdframed} 

\usepackage{lipsum}             
\usepackage{tocloft}            
\usepackage{afterpage}          
\usepackage{bbding}             
\usepackage{epigraph}           
\usepackage{minitoc}            
\usepackage{multicol}           
\usepackage{textgreek}          
\usepackage{graphicx,calc}
\usepackage{threeparttable}
\usepackage{placeins}

\usepackage{tcolorbox}
\tcbuselibrary{breakable}

\newcolumntype{P}[1]{>{\RaggedRight\arraybackslash}p{#1}}

\newcolumntype{C}{>{\centering\arraybackslash}X}

\definecolor{mygray}{gray}{0.92}
\definecolor{uclablue}{RGB}{39, 116, 174}
\definecolor{bigaired}{RGB}{156, 0, 0}
\definecolor{myblue}{HTML}{598BE7}
\definecolor{mildblue}{RGB}{31,119,180}
\definecolor{sectionblue}{RGB}{70, 130, 180}
\definecolor{methodblue}{RGB}{0, 150, 136}
\definecolor{bgblue}{RGB}{245,243,253}
\definecolor{ttblue}{RGB}{91,194,224}
\definecolor{mygreen}{rgb}{0.64, 0.56, 0.88}
\definecolor{myyellow}{rgb}{0.68, 0.6, 0.1}
\definecolor{fancygreen}{rgb}{0.33, 0.68, 0.20}
\definecolor{salmon}{rgb}{0.94, 0.52, 0.49}
\definecolor{tablegreen}{rgb}{0.82, 0.94, 0.75}
\definecolor{tableblue}{rgb}{0.81, 0.90, 0.94}
\definecolor{tablered}{rgb}{0.97, 0.85, 0.85}
\definecolor{tableorange}{rgb}{0.96, 0.85, 0.81}
\definecolor{myorange}{rgb}{1.0, 0.49, 0.0}
\definecolor{tlgreen}{rgb}{0.33, 0.68, 0.20}
\definecolor{darkgreen}{RGB}{0,100,0}
\definecolor{darkred}{RGB}{200, 0, 0}
\definecolor{customyellow}{HTML}{FFFACD}
\definecolor{refinegreen}{RGB}{0, 128, 75}
\definecolor{scoregreen}{RGB}{34, 139, 34}
\definecolor{hidden-blue}{RGB}{194,232,247}
\definecolor{hidden-black}{RGB}{20,68,106}
\definecolor{yes}{HTML}{C6EFCE}
\definecolor{no}{HTML}{FFC7CE}
\definecolor{partial}{HTML}{FFEB9C}
\definecolor{external}{HTML}{D9E1F2}
\definecolor{hdr}{HTML}{F2F2F2}
\definecolor{GRPOrow}{gray}{0.96}
\definecolor{FlowRLrow}{RGB}{225,236,255}
\definecolor{FlowBlue}{RGB}{80,120,210}
\definecolor{GRPOGray}{gray}{0.35}

\definecolor{gptgray}{RGB}{245,246,248}
\definecolor{gptbar}{RGB}{230,232,236}
\definecolor{qwenlav}{RGB}{244,240,252}
\definecolor{qwenbar}{RGB}{228,220,246}

\definecolor{myGreen}{RGB}{0, 150, 0}
\definecolor{myRed}{RGB}{200, 0, 0}

\setlist[itemize]{leftmargin=20pt, noitemsep, topsep=0pt}

\NewDocumentCommand{\kaiyan}{mO{}}{\textcolor{purple}{\textsuperscript{\textit{kaiyan}}\textsf{\textbf{\small[#1]}}}}
\NewDocumentCommand{\yuxin}{mO{}}{\textcolor{cyan}{\textsuperscript{\textit{yuxin}}\textsf{\textbf{\small[#1]}}}}
\NewDocumentCommand{\bx}{mO{}}{\textcolor{green}{\textsuperscript{\textit{bx}}\textsf{\textbf{\small[#1]}}}}
\NewDocumentCommand{\at}{mO{}}{\textcolor{red}{\textsuperscript{\textit{AT}}\textsf{\textbf{\small[#1]}}}}
\NewDocumentCommand{\re}{mO{}}{\textcolor{blue}{\textsuperscript{\textit{RE}}\textsf{\textbf{\small[#1]}}}}
\NewDocumentCommand{\ybsun}{mO{}}{\textcolor{magenta}{\textsuperscript{\textit{youbang}}\textsf{\textbf{\small[#1]}}}}
\NewDocumentCommand{\runze}{mO{}}{\textcolor{orange}{\textsuperscript{\textit{runze}}\textsf{\textbf{\small[#1]}}}}
\NewDocumentCommand{\add}{mO{}}{\textcolor{darkgreen}{\textsuperscript{\textit{Maybe Consider Discuss}}\textsf{\textbf{[#1]}}}}

\newcommand{\cmark}{\textcolor{darkgreen}{\boldmath$\checkmark$}}
\newcommand{\xmark}{\textcolor{darkred}{\boldmath$\times$}}

\newenvironment{itemize*}%
 {\leftmargini=10pt\begin{itemize}%
  \setlength{\itemsep}{0pt}%
  \setlength{\parskip}{0pt}%
  }%
 {\end{itemize}}

\newenvironment{enumerate*}%
 {\begin{enumerate}%
  \setlength{\itemsep}{0pt}%
  \setlength{\parskip}{0pt}}%
 {\end{enumerate}}

\newcommand{\cellstatus}[1]{%
  \begingroup
  \StrTrim{#1}[\statusval]%
  \IfStrEq{\statusval}{Yes}{\cellcolor{yes}\cmark}{}%
  \IfStrEq{\statusval}{No}{\cellcolor{no}\xmark}{}%
  \IfBeginWith{\statusval}{Yes (}{\cellcolor{yes}\cmark~\textit{\statusval\unskip}}{}%
  \IfStrEq{\statusval}{Partial}{\cellcolor{partial}\textbf{Partial}}{}%
  \IfStrEq{\statusval}{External}{\cellcolor{external}\textbf{External}}{}%
  \endgroup
}

\newtcolorbox{myboxi}[1][]{
  breakable,
  title=#1,
  colback=red!5,
  colbacktitle=red!5,
  coltitle=black,
  fonttitle=\bfseries,
  bottomrule=0pt,
  toprule=0pt,
  leftrule=2pt,
  rightrule=2pt,
  titlerule=0pt,
  arc=0pt,
  outer arc=0pt,
  colframe=red,
}

\newtcolorbox{myboxnote}[1][]{
  breakable,
  title=#1,
  colback=orange!0,
  colbacktitle=orange!0,
  coltitle=black,
  fonttitle=\bfseries,
  bottomrule=0pt,
  toprule=0pt,
  leftrule=2pt,
  rightrule=2pt,
  titlerule=0pt,
  arc=0pt,
  outer arc=0pt,
  colframe=orange,
}

\newtcolorbox{myboxii}[1][]{
  breakable,
  freelance,
  title=#1,
  colback=white,
  colbacktitle=white,
  coltitle=black,
  fonttitle=\bfseries,
  bottomrule=0pt,
  boxrule=0pt,
  colframe=white,
  overlay unbroken and first={
  \draw[red!75!black,line width=3pt]
    ([xshift=5pt]frame.north west) -- 
    (frame.north west) -- 
    (frame.south west);
  \draw[red!75!black,line width=3pt]
    ([xshift=-5pt]frame.north east) -- 
    (frame.north east) -- 
    (frame.south east);
  },
  overlay unbroken app={
  \draw[red!75!black,line width=3pt,line cap=rect]
    (frame.south west) -- 
    ([xshift=5pt]frame.south west);
  \draw[red!75!black,line width=3pt,line cap=rect]
    (frame.south east) -- 
    ([xshift=-5pt]frame.south east);
  },
  overlay middle and last={
  \draw[red!75!black,line width=3pt]
    (frame.north west) -- 
    (frame.south west);
  \draw[red!75!black,line width=3pt]
    (frame.north east) -- 
    (frame.south east);
  },
  overlay last app={
  \draw[red!75!black,line width=3pt,line cap=rect]
    (frame.south west) --
    ([xshift=5pt]frame.south west);
  \draw[red!75!black,line width=3pt,line cap=rect]
    (frame.south east) --
    ([xshift=-5pt]frame.south east);
  },
}

\tcbset{
  takeawaysbox/.style={
    title=Takeaways,
    colback=lightblue!80,
    colframe=black,
    fonttitle=\bfseries\small,
    coltitle=white,
    colbacktitle=black,
    enhanced,
    attach boxed title to top left={xshift=2.5mm,yshift=-2.5mm},
    boxed title style={rounded corners, size=small, colframe=black, colback=black},
    width=\linewidth,
    arc=3.5mm
  }
}

\mdfdefinestyle{mystyle}{%
  rightline=true,
  innerleftmargin=10,
  innerrightmargin=10,
  outerlinewidth=3pt,
  topline=false,
  rightline=true,
  bottomline=false,
  skipabove=\topsep,
  skipbelow=\topsep
}

\tikzset{%
    every node/.style={font=\tiny},
    parent/.style =          {align=center,text width=2cm,rounded corners=3pt, line width=0.3mm, fill=gray!10,draw=gray!80},
    child/.style =           {align=center,text width=2.0cm,rounded corners=3pt, fill=blue!10,draw=blue!80,line width=0.3mm},
    grandchild/.style =      {align=center,text width=2cm,rounded corners=3pt},
    greatgrandchild/.style = {align=center,text width=1.5cm,rounded corners=3pt},
    greatgrandchild2/.style = {align=center,text width=1.5cm,rounded corners=3pt},    
    referenceblock/.style =  {align=center,text width=1.5cm,rounded corners=2pt},
    pretrain/.style =           {align=center,text width=2.0cm,rounded corners=3pt, fill=blue!10,draw=blue!80,line width=0.3mm},   
    pretrain_work/.style =           {align=center, text width=8.5cm,rounded corners=3pt, fill=blue!10,draw=blue!0,line width=0.3mm},  
    template/.style =           {align=center,text width=2.0cm,rounded corners=3pt, fill=red!10,draw=red!80,line width=0.3mm},   
    template_work/.style =           {align=center,text width=8.5cm,rounded corners=3pt, fill=red!10,draw=red!0,line width=0.3mm},    
    answer/.style =           {align=center,text width=2.0cm,rounded corners=3pt, fill= cyan!10,draw= cyan!80,line width=0.3mm},   
    answer_work/.style =           {align=center,text width=8.5cm,rounded corners=3pt, fill= cyan!10,draw= cyan!0,line width=0.3mm},      
    multiple/.style =           {align=center,text width=2.0cm,rounded corners=3pt, fill= orange!10,draw= orange!80,line width=0.3mm},   
    multiple_work/.style =           {align=center,text width=8.5cm,rounded corners=3pt, fill= orange!10,draw= orange!0,line width=0.3mm},        
    tuning/.style =           {align=center,text width=2.0cm,rounded corners=3pt, fill= magenta!10,draw= magenta!80,line width=0.3mm},   
    tuning_work/.style =           {align=center,text width=8.5cm,rounded corners=3pt, fill= magenta!10,draw= magenta!0,line width=0.3mm},          
}

\newcommand{\lstbg}[3][0pt]{{\fboxsep#1\colorbox{#2}{\strut #3}}}

\lstdefinelanguage{diff}{
  basicstyle=\ttfamily\small,
  morecomment=[f][\lstbg{red!20}]-,
  morecomment=[f][\lstbg{green!20}]+,
}

\lstdefinelanguage{diffpython}{
  language=diff,
  morekeywords={def, if, else, for, while, return, import, from, as, class, with, try, except, finally, raise, lambda, and, or, not, in, is, None, True, False},
  morecomment=[l]{\#},
  morestring=[b]",
  morestring=[b]',
}

\usepackage{xspace}

\usepackage{xcolor}

\definecolor{ForestGreen}{RGB}{34,139,34}
\definecolor{myyellow}{RGB}{181, 181, 27}

\usepackage{amsmath}
\usepackage{amssymb}
\usepackage{mathtools}
\usepackage{amsthm}
\usepackage{fontawesome5} 

\usepackage[capitalize,noabbrev]{cleveref}

\usepackage{multirow}
\usepackage{makecell}
\usepackage{enumerate}
\usepackage{enumitem}
\usepackage{pifont}

\definecolor{mygrey}{gray}{0.4}

\usepackage[ruled,algo2e,vlined]{algorithm2e}
\SetKwInOut{Input}{Input}\SetKwInOut{Output}{Output}
\SetKwComment{Comment}{$\triangleright$\ }{}

\definecolor{darkgreen}{RGB}{30, 130, 30}
\definecolor{cream}{RGB}{253, 250, 242}
\definecolor{persona}{rgb}{0.376,0.741,1.0}
\renewcommand{\cmark}{\textcolor{darkgreen}{\ding{51}}} 
\renewcommand{\xmark}{\textcolor{red}{\ding{55}}}       

\newcommand{\ours}{\textsc{MobileMem}\xspace}

\usepackage{listings}

\usepackage{ulem}
\usepackage{pifont}

\setheadertext{Preprint}

\newcommand{\KEME}[0]{\textsc{KEME}\xspace}
\newcommand{\bench}[0]{\textsc{MobileMem}\xspace}

\newcommand{\omni}[0]{\textsc{MobileMem-Omni}\xspace}

\usepackage[most]{tcolorbox}
\usepackage{xcolor}

\definecolor{RemarkBlue}{RGB}{38,96,186}

\newcounter{assump}[section]

\newtcolorbox{remarkbox}{
    breakable,
    colback=RemarkBlue!5,
    colframe=RemarkBlue,
    boxrule=1.5pt,
    arc=2.5mm,
}

\title{\ours: Learning from a Year of Mobile Experiences}
\setheadertitle{\ours: On-Device Memory for Continually Evolving Agents}

\author{
OPPO, OpenKG\footnote{Corresponding Author: Ningyu Zhang (zhangningyu@zju.edu.cn)}
}

\githublink{https://github.com/zjunlp/MobileMem}

\begin{document}

\fancypagestyle{firststyle}{
    \fancyhead[L]{\includegraphics[height=20pt,keepaspectratio]{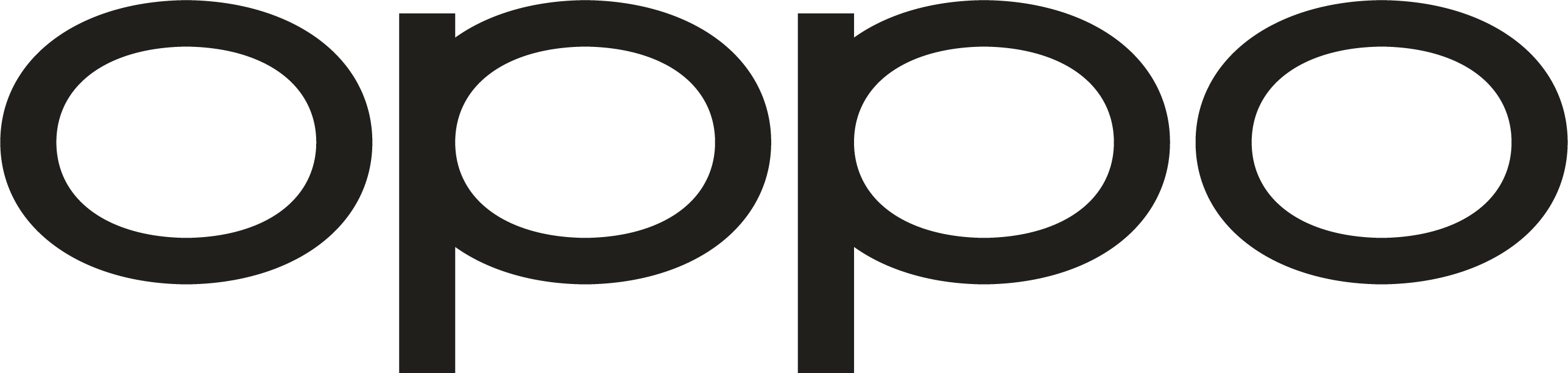}\hspace{10pt}\includegraphics[height=24pt,keepaspectratio]{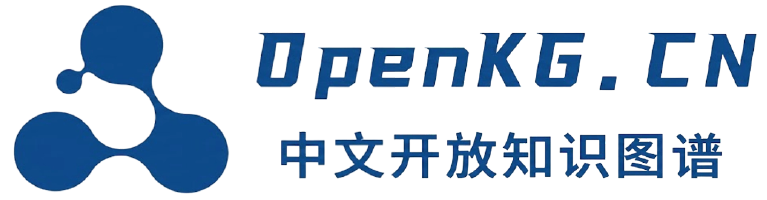}}
    \fancyhead[C]{}
    \fancyhead[R]{{\color{black}\normalfont\itshape\fontsize{9}{10}\selectfont Technical Report}}
    \renewcommand{\headrulewidth}{0pt}
    \fancyfoot[L]{}
    \fancyfoot[R]{}
    \fancyfoot[C]{}
}

\fancypagestyle{plain}{
    \fancyhead[L]{\includegraphics[height=20pt,keepaspectratio]{figures/oppo_logo.png}\hspace{10pt}\includegraphics[height=24pt,keepaspectratio]{figures/openkg_logo.pdf}}
    \fancyhead[C]{}
    \fancyhead[R]{{\color{black}\normalfont\itshape\fontsize{9}{10}\selectfont Technical Report}}
    \renewcommand{\headrulewidth}{0.4pt}
    \fancyfoot[L]{}
    \fancyfoot[R]{}
    \fancyfoot[C]{}
}

\fancyhead[L]{{\color{black}\normalfont\itshape\fontsize{9}{10}\selectfont Technical Report}}
\fancyhead[C]{}
\fancyhead[R]{\includegraphics[height=20pt,keepaspectratio]{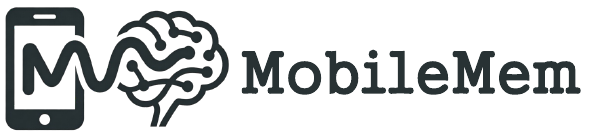}}
\fancyfoot[L]{}
\fancyfoot[R]{}
\fancyfoot[C]{\thepage}
\renewcommand{\headrulewidth}{0.4pt}
\setheadertitle{MobileMem: On-Device Memory for Continually Evolving Agents}

\begin{abstract}
The next generation of AI agents is increasingly moving beyond systems that answer isolated questions toward persistent personal assistants that can understand, remember, and continuously learn from users’ experiences. Such assistants require long-term memory to accumulate and leverage user-specific experiences over time, yet existing benchmarks remain inadequate for realistic mobile settings, where experiences are heterogeneous, multimodal, evolving, and deeply personal. We introduce \ours, a benchmark and framework for studying on-device long-term memory, grounded in a year-scale collection of mobile experiences. \ours employs a knowledge-grounded synthesis pipeline to construct coherent and temporally consistent long-horizon trajectories from user-app sessions. It provides complementary text and multimodal settings covering multi-hop and temporal reasoning, knowledge updating, and implicit preference inference. Specifically, \ours enables agents to remember the past, understand the present, and adapt to the future. By modeling experiences rather than isolated facts, \ours moves memory beyond information retrieval toward experiential intelligence for continuous personal learning.

\end{abstract}


\pagenumbering{gobble}
\maketitle

\vspace{1em}
\begin{center}
    \nopagebreak 
    \includegraphics[width=\textwidth, keepaspectratio]{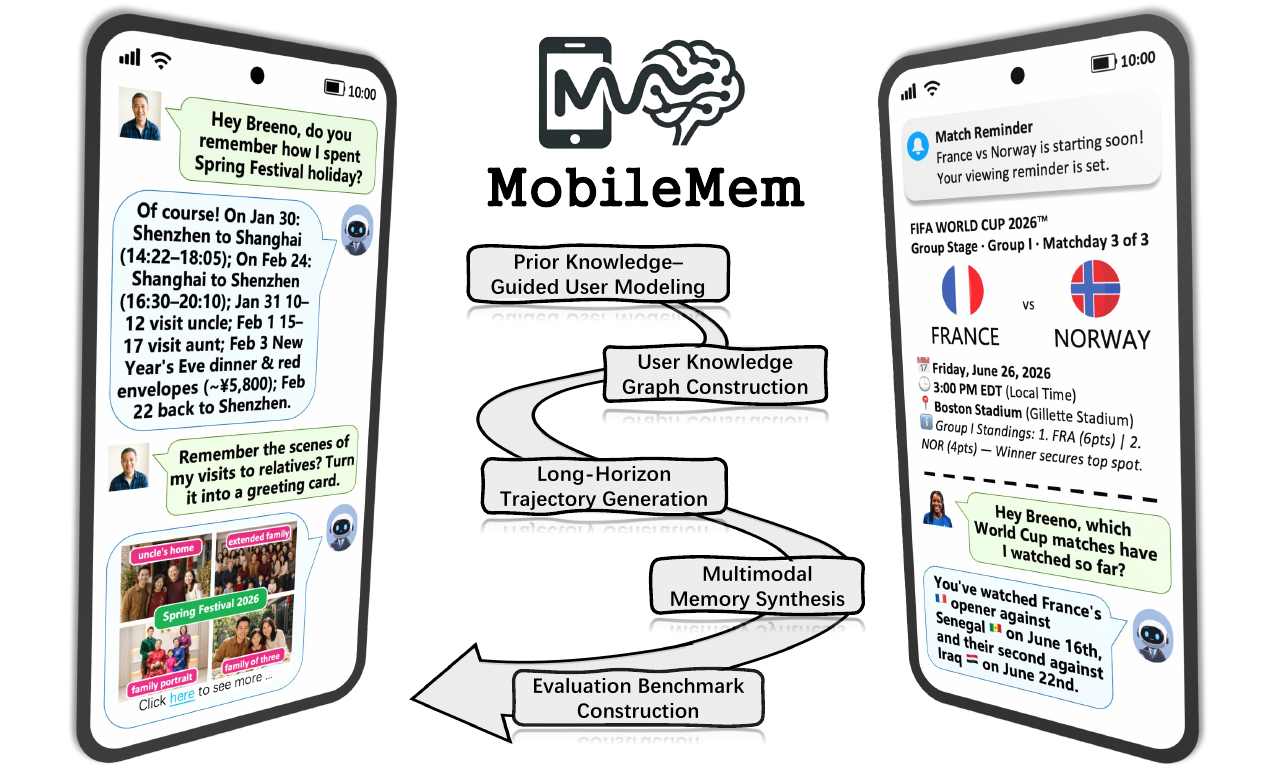}
    \captionof{figure}{The overview of \ours.}
    \label{fig:mobilemem_overview}
\end{center}
\vspace{1em}

\renewcommand{\contentsname}{Contents}
\tableofcontents

\clearpage
\pagenumbering{arabic}
\section{Introduction}

Over the past decade, advances in Large Language Models (LLMs) \citep{llm_survey1,llm_survey2,mllm_survey1} have substantially enhanced AI systems’ capabilities in knowledge acquisition, reasoning and tool use, enabling agents to move beyond static question answering toward increasingly capable and interactive systems. 
As these capabilities continue to mature, AI agents are moving toward a new paradigm of \textbf{persistent personal intelligence}, in which agents accompany users across days, months, and even years rather than serving as isolated question-answering systems. 
This paradigm is emerging across diverse forms, including AI phones, AI glasses, personal copilots, embodied assistants, and automotive agents \citep{ColorAgent, yao2026joyai,chen2026lightmem, DBLP:journals/corr/abs-2509-26536, DBLP:journals/corr/abs-2606-31830, WorldMM, realworldhsimulation, PersonalizedDeepResearch, DBLP:journals/corr/abs-2607-08448}. 
These agents are expected not only to understand and respond to users in the moment, but also to remember past experiences, understand evolving contexts, learn from everyday interactions, and continuously adapt to their users. 
Their intelligence will therefore be defined not solely by what they know, but increasingly by \textbf{what they remember, what they learn from experience, and how well they evolve with their users}.

Realizing such persistent personal intelligence requires agents to maintain a \textbf{continuous understanding of their users} rather than treating each interaction as an isolated task.
This understanding must be grounded not only in the current context, but also in users’ past experiences, evolving preferences, routines, relationships, and behaviors.
To this end, growing efforts have explored personalization \citep{ColorAgent, ClawMobile, Mi-Memory} and the development of personalized LLMs \citep{chen2024large, liu2025survey}, personalized MLLMs \citep{YoLLaVA, PersonaVLM}, and personalized AI agents \citep{Xu2026TowardPL}.
However, existing personalization approaches often focus primarily on adapting models or agents to the user at a given moment, while persistent personal intelligence requires them to \textbf{accumulate, organize, and learn from experiences over extended periods}.
As interactions unfold, agents must transform fragmented observations into persistent knowledge about the user, while continuously updating this knowledge as the user’s life and preferences evolve.
This makes long-term memory \citep{survey1, survey2, survey3, survey4,yu2026agentic, guo2026memfactory} a foundational capability for persistent personal AI, enabling agents to learn from the past, contextualize the present, and continuously adapt to the future.

\textbf{Mobile Memory.} Existing research \citep{Zep, MemGPT, Mem0, AMEM, MemOS, Mirix, EverMemOS, Memory-R1, yao2025rethinking,M2A, DBLP:conf/acl/LuLSWWH26, DBLP:conf/gpce/KottenhahnK26}  on agent memory largely operates under assumptions of infinite storage, cloud-based computation, and centralized databases, which are characteristic of cloud-based scenarios \citep{survey5, DBLP:conf/acl/ChenLZHZ26, DBLP:conf/aaai/SunLHZ26, DBLP:conf/acl/LiLLHZ26, DBLP:conf/aaai/LiSHH025}. However, the primary deployment target for next-generation agents \citep{M3, EgoLife, Mem-Gallery, MemEye, MemLens, ATMbench, WorldMM} is not the cloud but mobile devices, including smartphones, AR glasses, smart earbuds, wearables, and in-car assistants. These devices naturally generate first-person video streams, audio feeds, user interaction logs, geolocation trajectories, and multimodal interaction histories, collectively forming rich personal digital memories. This shift raises a fundamental research question: \textit{Can an agent build and evolve its memory entirely on-device?} 
Unlike stationary devices, mobile devices are portable and easy to use anywhere, which leads them to continuously generate large amounts of highly personal data every day, including chat logs, photos, app screenshots, social media content, and shopping information. Such data are inherently multimodal, cross-app, and strongly temporally continuous. Important user memories often take the form of events distributed across different modalities and applications. Therefore, on-device memory requires continuous modeling of the user's multimodal event stream and app interactions, marking a paradigm shift from cloud-based memory to mobile memory.
As foundation models become increasingly homogenized, the true differentiator is no longer the model itself, but whether it understands the person it serves. The ability to build accurate, secure, and durable on-device personal memory systems will determine who dominates the next generation of intelligent services. Behind this lies an accelerating trend: every individual will have their own AI, and every AI will need its own on-device memory. No longer a replaceable feature, on-device memory is the foundation upon which intelligent agents operate. Just as cloud computing and databases constitute today's digital infrastructure, on-device memory will become a new critical factor of production in the digital economy era.

\begin{figure}[t]
    \centering
    \includegraphics[width=1\columnwidth]{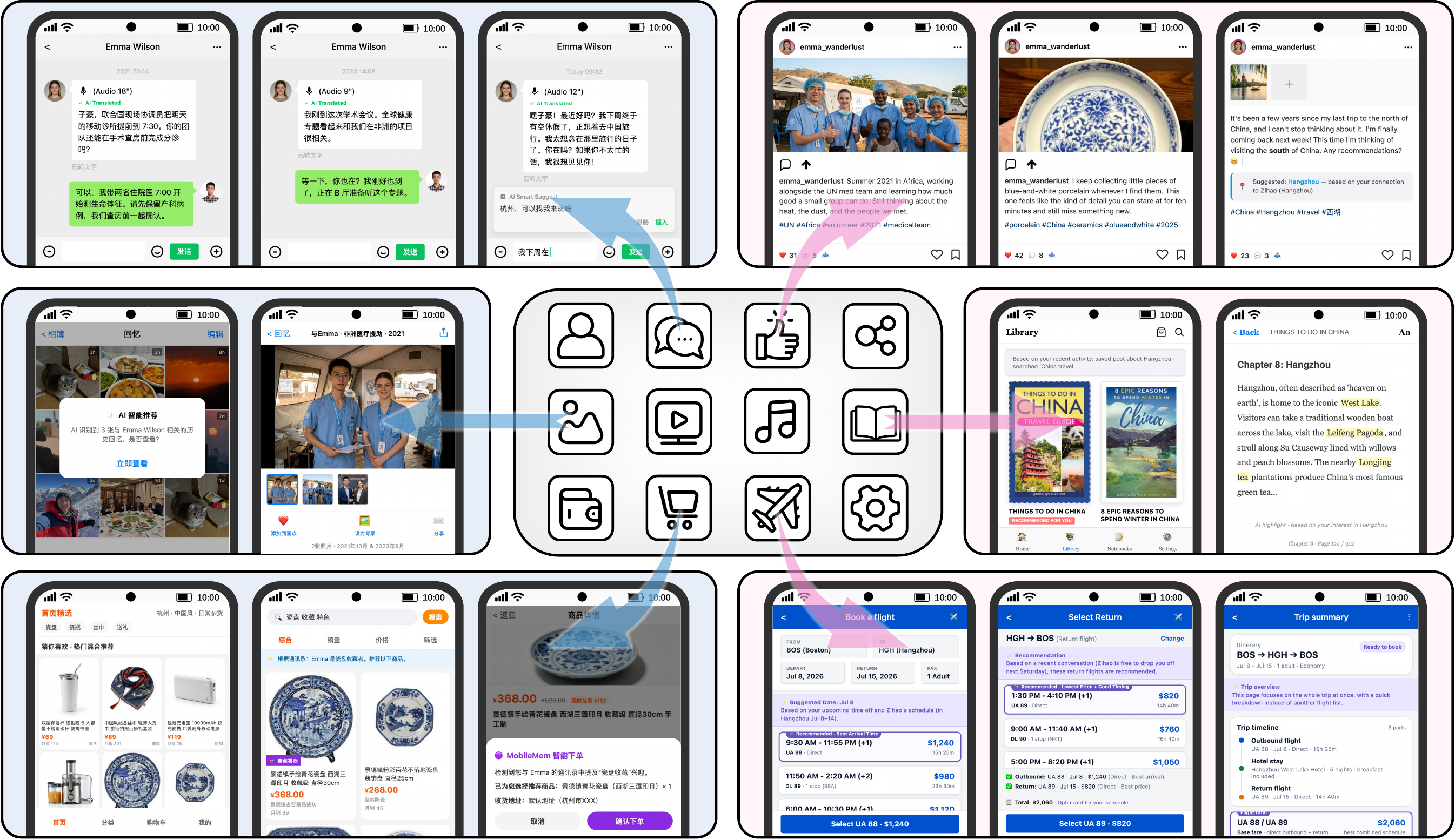}
    \caption{
    \ours breaks down data silos across chat, photos, reading, shopping, and travel apps, unifying fragmented data into one memory layer for proactive personalization.
    }
    \label{fig:data_silos_unification}
    \vspace{-1em}
\end{figure}

\textbf{Challenges of Evolving Mobile Memory.}
The on-device environment differs fundamentally from the cloud, introducing a distinct set of challenges for building continually evolving memory systems. First, mobile devices face inevitable memory explosion as users generate thousands of events, hundreds of images, long-duration videos, and continuous voice interactions daily, overwhelming limited on-device storage and bandwidth. Second, memories are inherently fragmented across modalities and applications, lacking a unified semantic structure that makes it difficult for agents to construct coherent holistic memories from disparate sources. Third, user interests, habits, and contexts continuously evolve, requiring memory systems to dynamically adapt by identifying stale information and incorporating new experiences without catastrophic forgetting. Fourth, mobile devices operate under severe constraints on storage, power consumption, latency, and computational capacity, making it ineffective to directly port cloud-based memory solutions. Fifth, personal memories contain highly sensitive information, and uploading such data to the cloud exposes significant privacy risks, necessitating local-first processing. Therefore, memory must be synthesized, organized, and utilized locally, marking a fundamental paradigm shift from cloud-centric to on-device memory systems.

\textbf{The \bench Benchmark.}
Existing long-term memory benchmarks \citep{LoCoMo, Mem-Gallery, HaluMem, PersonaMem2, LongMemEval} fail to capture the heterogeneous, multimodal, and continuously evolving nature of mobile interactions, making them inadequate for evaluating on-device memory systems in realistic mobile scenarios.
As illustrated in Figure~\ref{fig:mobilemem_overview},  we propose \bench, a comprehensive benchmarking framework for evaluating on-device memory systems.
\bench is specifically designed to assess the core capabilities of mobile memory architecture, which captures the full lifecycle of user experiences through structured personal knowledge graphs that represent entities, relations, and temporal dynamics across modalities and applications. 
It evaluates how effectively memory systems transform raw fragmented events into coherent episodic and semantic representations, and adapt to evolving user contexts without catastrophic forgetting.
Through systematic evaluation across multiple reasoning categories and realistic mobile scenarios, \bench provides standardized metrics for long-horizon memory preservation and individual knowledge accumulation, thereby establishing a foundation for advancing the next generation of intelligent, evolving personal assistants on mobile devices.

\textbf{Long-Horizon Data Synthesis from User Prior Knowledge.}
Realistic evaluation of long-term memory requires interaction data that captures not only what users know, but also how their experiences unfold and evolve over time. 
However, collecting such long-horizon, personalized, and multimodal trajectories is costly and raises privacy and data-availability challenges. 
We therefore propose a knowledge-grounded synthesis framework \KEME that constructs evolving user experiences from structured prior knowledge, using personas, temporal constraints, and existing interactions as knowledge anchors to generate coherent long-term trajectories. 
Built on this idea, \KEME alternates between top-down temporal planning and bottom-up experience evolution to synthesize session-level interactions that are temporally consistent and reflect changing user states, while also generating multi-hop QA pairs that require reasoning across events and time. 

Based on this framework, \bench instantiates two complementary benchmark scenarios. 
\bench targets textual settings where applications integrate with the system-level assistant via predefined templates, evaluating cross-session information retrieval and logical inference from dialogue logs and application interactions across reasoning categories including multi-hop reasoning, temporal reasoning, knowledge updating, implicit preference inference, and abstention. 
\omni extends evaluation to multimodal settings where applications operate independently from the memory layer, requiring users to share screenshots of important interactions with the assistant. 
This scenario introduces additional challenges in visual reasoning and cross-modal retrieval, incorporating real-user participation, synthetic private images, and bilingual role-based interactions. 
Together, these benchmarks provide a comprehensive framework for assessing how effectively memory systems synthesize, organize, and retrieve structured knowledge from raw mobile experiences.

\section{Preliminaries}

\subsection{Problem Formulation}
As shown in Figure~\ref{fig:data_silos_unification}, we consider an on-device personal assistant scenario, where a core mobile assistant, such as Breeno from OPPO or Siri from Apple, runs on a user's smartphone and provides personalized services via an associated memory layer.

To evaluate such memory layers, each instance in the \bench-series benchmarks is formalized as a tuple $(\mathcal{T}, \mathcal{Q})$, where $\mathcal{T}$ denotes the interaction trajectory and $\mathcal{Q} = \{(q_j, \mathcal{A}_j)\}_{j=1}^{|\mathcal{Q}|}$ denotes the set of question-answer pairs associated with the trajectory. Here, $\mathcal{A}_j = \{a_k^{(j)}\}_{k=1}^{|\mathcal{A}_j|}$ is the set of golden answers for question $q_j$. Unlike prior conversational memory benchmarks that mainly focus on human-assistant dialogues, \ours models the interaction trajectory as \textit{a stream of heterogeneous actions}. These actions may be initiated by the user or the assistant. More importantly, in realistic mobile environments, both the user and the assistant interact not only with each other but also with a wide range of external applications. The involvement of third-party applications makes the action stream highly heterogeneous and potentially multi-modal, including natural language utterances, application operations, interface states, notifications, and other application-specific interaction records. Typically, this continuous action stream is segmented into discrete sessions $\mathcal{T} = \{\mathcal{S}_j\}_{j=1}^{|\mathcal{T}|}$ according to a predefined segmentation criterion \citep{DBLP:journals/csur/WangCWSOL22, DBLP:conf/acl/XuSW22, Wang2025OPeRAAD}, where $\mathcal{S}_j = \{x_k^{(j)}\}_{k=1}^{|\mathcal{S}_j|}$ represents the $j$-th session. For notational simplicity, we flatten the session-level trajectory into a single action sequence $\tau = \operatorname{Flatten}(\mathcal{T}) = \{x_i\}_{i=1}^{|\tau|}$, where $\operatorname{Flatten}(\cdot)$ concatenates all sessions in chronological order. Let $\mathcal{M}_0$ denote the initial memory state. At step $t$ ($t \geq 1$), the memory layer receives action $x_t$ and updates its memory state to $\mathcal{M}_t$ via the update function:
$
\mathcal{M}_t = \operatorname{Update}(\mathcal{M}_{t-1}, x_t),
$
where $\operatorname{Update}(\cdot)$ denotes the memory update mechanism.

We adopt an end-to-end strategy to evaluate the memory system. Specifically, for each question-answer pair $(q, \mathcal{A}) \in \mathcal{Q}$, the memory system generates an answer $\hat{a}$ based on the final memory state $\mathcal{M}_{|\tau|}$, which integrates information from the entire interaction trajectory $\tau$. The correctness of $\hat{a}$ is assessed against the reference answer set $\mathcal{A}$ via an LLM-as-a-Judge protocol \citep{Gu2024ASO, LLM-as-a-judge}.

\subsection{On-Device Memory Layer}

\begin{figure}[t]
    \centering
    \includegraphics[width=1\columnwidth]{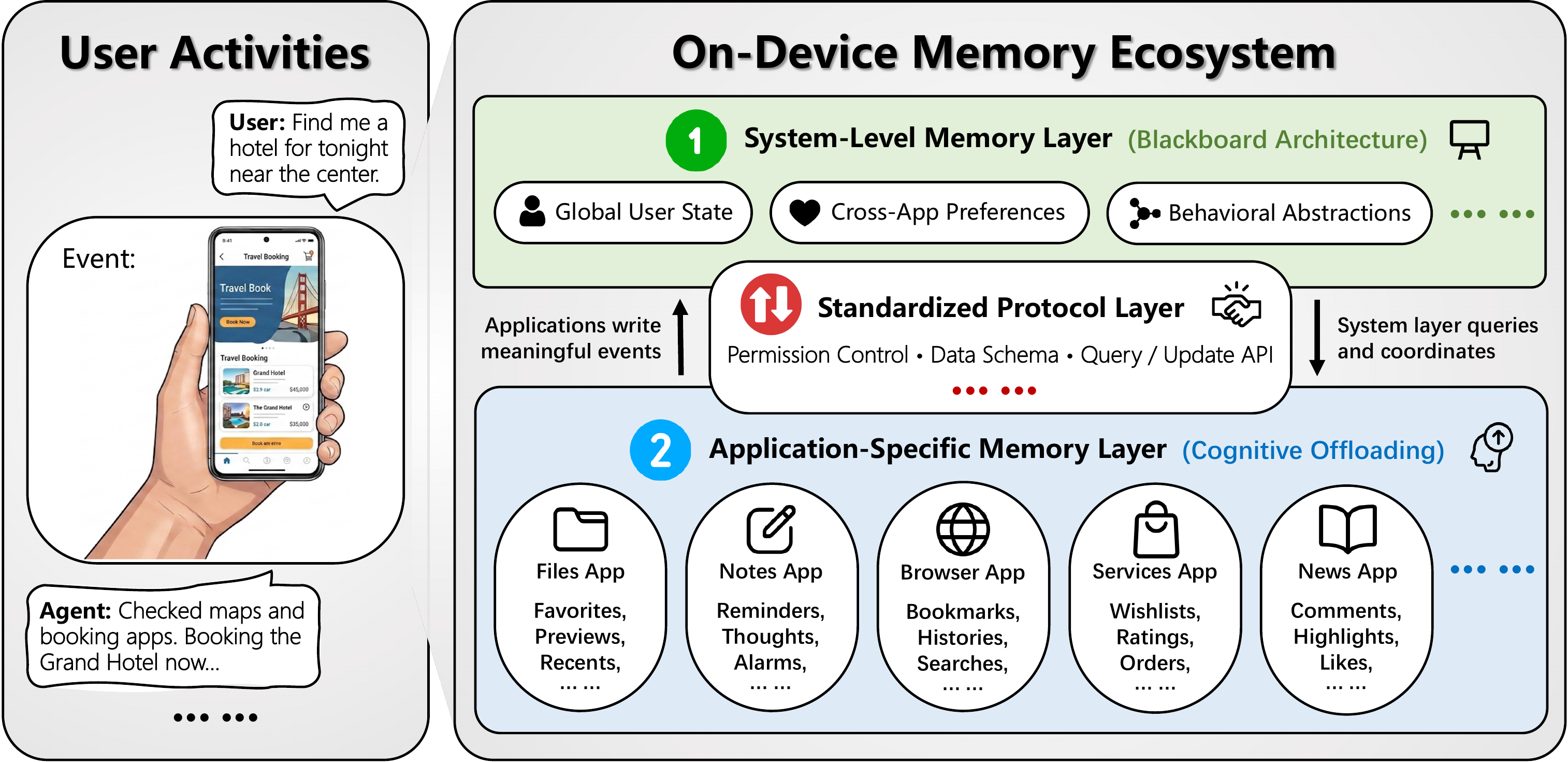}
    \caption{
    The position of \ours in the envisioned on-device memory ecosystem, with a two‑layer composite architecture and an information exchange protocol between them.
    }
    \label{fig:memory_ecosystem_architecture}
    \vspace{-1em}
\end{figure}

Existing memory layers are primarily designed for conversational settings, where the interaction history consists mainly of user-assistant dialogues. However, \textbf{in mobile environments, a substantial portion of user activities occurs outside the assistant through interactions with third-party applications}, such as browsing webpages, reading books, shopping online, or managing personal finances. These activities are often only \textit{partially observable} to the memory layer.

To address this \textit{observability gap}, recent systems like MRIXI \citep{Mirix} and MineContext \citep{MineContext2025} adopt screenshot-based perception mechanisms, periodically capturing device screens and inferring user activities from visual changes across consecutive screenshots. While this strategy expands the observable interaction space, it treats applications as passive observation targets and requires the memory layer to reconstruct high-level user behaviors from noisy interface-level signals, introducing redundancy, privacy concerns, and substantial processing overhead. \textbf{We argue that this perspective overlooks an important property of modern mobile ecosystems: many applications already function as specialized memory systems}. Applications routinely store historical user interactions in local or cloud databases and expose retrieval mechanisms over these records. For example, a web browser maintains browsing histories with associated timestamps and supports keyword-based retrieval. Conceptually, this resembles memory systems that organize historical experiences into searchable memory stores and retrieve them through lexical matching mechanisms such as BM25. Similarly, financial management applications persist transaction records and continuously aggregate them into higher-level statistics, such as monthly income and expenditure summaries. Such aggregation behaviors are analogous to memory consolidation processes that transform raw experiences into more compact and semantically meaningful representations. Many applications further provide structured retrieval capabilities, including temporal filtering, categorical search, and summary generation. 

From this perspective, on-device memory should not be viewed as a single monolithic memory store. \textbf{Instead, we envision a composite memory architecture consisting of two complementary components}. As shown in Figure~\ref{fig:memory_ecosystem_architecture}, the first component is a general-purpose memory layer associated with the system-level assistant (e.g., Breeno or Siri), which maintains cross-application user knowledge and supports personalized assistance. The second component comprises application-specific memory systems, where each application manages, organizes, and retrieves information within its own domain. These two components can collaborate through \textit{standardized protocols}, forming a unified memory ecosystem whose capability exceeds the simple aggregation of individual modules.

\paragraph{System-Level Memory Layer as a Blackboard.} From the perspective of inter-component coordination, the system-level memory layer plays a role analogous to a \textit{blackboard architecture} \citep{DBLP:journals/ai/Hayes-Roth85}. It serves as a shared workspace that stores global user state, cross-application preferences, and high-level behavioral abstractions that are useful for future reasoning and task execution. In this analogy, application-specific memory systems act as \textit{knowledge sources}, each specializing in a distinct domain. The personal assistant functions as a \textit{controller}, responsible for solving user tasks. This design enables a bidirectional flow of information. Applications can write meaningful behavioral records into the system-level memory layer. For example, a finance app may continuously record transaction events and forward these behavioral traces to the system-level memory layer through the standardized interface. By aggregating such signals over time, the system-level memory layer can extract higher-level behavioral patterns, such as a user's tendency to purchase food from barbecue restaurants late at night, enabling the assistant to provide more personalized services. Conversely, the system-level memory layer can leverage these accumulated abstractions to coordinate information retrieval across multiple application-specific memories. Given a user request, the assistant first reads the current system-level memory state and then searches memories from relevant applications through standardized interfaces. The retrieved information is subsequently integrated and reasoned over to support cross-application decision-making and personalized assistance.

\paragraph{Application-Specific Memory as Cognitive Offloading.} Application-specific memory systems serve as a form of \textit{cognitive offloading} \citep{Risko2016CognitiveO} for the system-level memory layer. Mobile environments are characterized by highly heterogeneous and fine-grained user interactions across diverse applications, making it infeasible for a centralized memory system to directly process all raw behavioral signals. During memory construction, applications can act as the perception layer of the memory ecosystem, performing an initial stage of memory extraction before information reaches the system-level memory. Rather than forwarding every low-level interaction, applications selectively expose events that are likely to contribute to higher-level user modeling. For example, when a user creates and saves a non-empty note, the note application may package the note content, title, and timestamp into a structured representation and forward it to the system-level memory layer. In contrast, transient interface operations such as scrolling, page switching, or navigation behaviors need not be propagated. This application-side filtering substantially reduces redundant signals and allows the system-level memory layer to focus on extracting long-term preferences and behavioral patterns. During memory retrieval, applications continue to serve as specialized memory repositories. When detailed historical information is required, the assistant can query the relevant application memories through standardized protocols and integrate the retrieved information into downstream reasoning. Overall, this division of responsibilities allows application-specific memories to absorb much of the storage, organization, and retrieval complexity associated with low-level user interactions. As a result, the system-level memory layer can concentrate on high-level abstraction, cross-application reasoning and personalization, leading to a more \textit{scalable} and more \textit{privacy-preserving} memory architecture.

Note that all interactions between the two components are governed by a standardized protocol that allows both users and developers to define the boundaries of information exchange. Users may specify which applications are permitted to contribute to system-level memory, while applications can determine which information is safe to expose, preventing unintended leakage of sensitive data.

\subsection{Benchmark Scenarios}
\label{sec:benchmark_scenarios}

The memory ecosystem described above represents a long-term vision for on-device memory systems. As an initial step toward evaluating memory capabilities in mobile environments, \bench and \omni instantiate two simplified yet representative scenarios.

\paragraph{\bench.} \bench models a setting where applications are tightly integrated with the system-level assistant through predefined templates. We focus on applications that naturally maintain user-generated records, such as note-taking, document editing, and voice memo applications. In this setting, user interactions are restricted to record creation operations.
Whenever a new record is created, the application extracts the corresponding content and forwards it to the system-level memory layer as a structured memory event. 

\paragraph{\omni.} \omni considers a more general setting where applications are not directly connected to the system-level memory layer. Instead, we assume that users can share screenshots of important application interactions with the assistant. These screenshots serve as an observation channel through which the assistant gains access to application-specific information. Compared with \bench, this setting introduces additional challenges in memory extraction as personalized information is distributed across both textual and visual modalities.

\section{Long-Horizon Data Synthesis from User Prior Knowledge}

\begin{figure}[t]
    \centering
    \includegraphics[width=1\columnwidth]{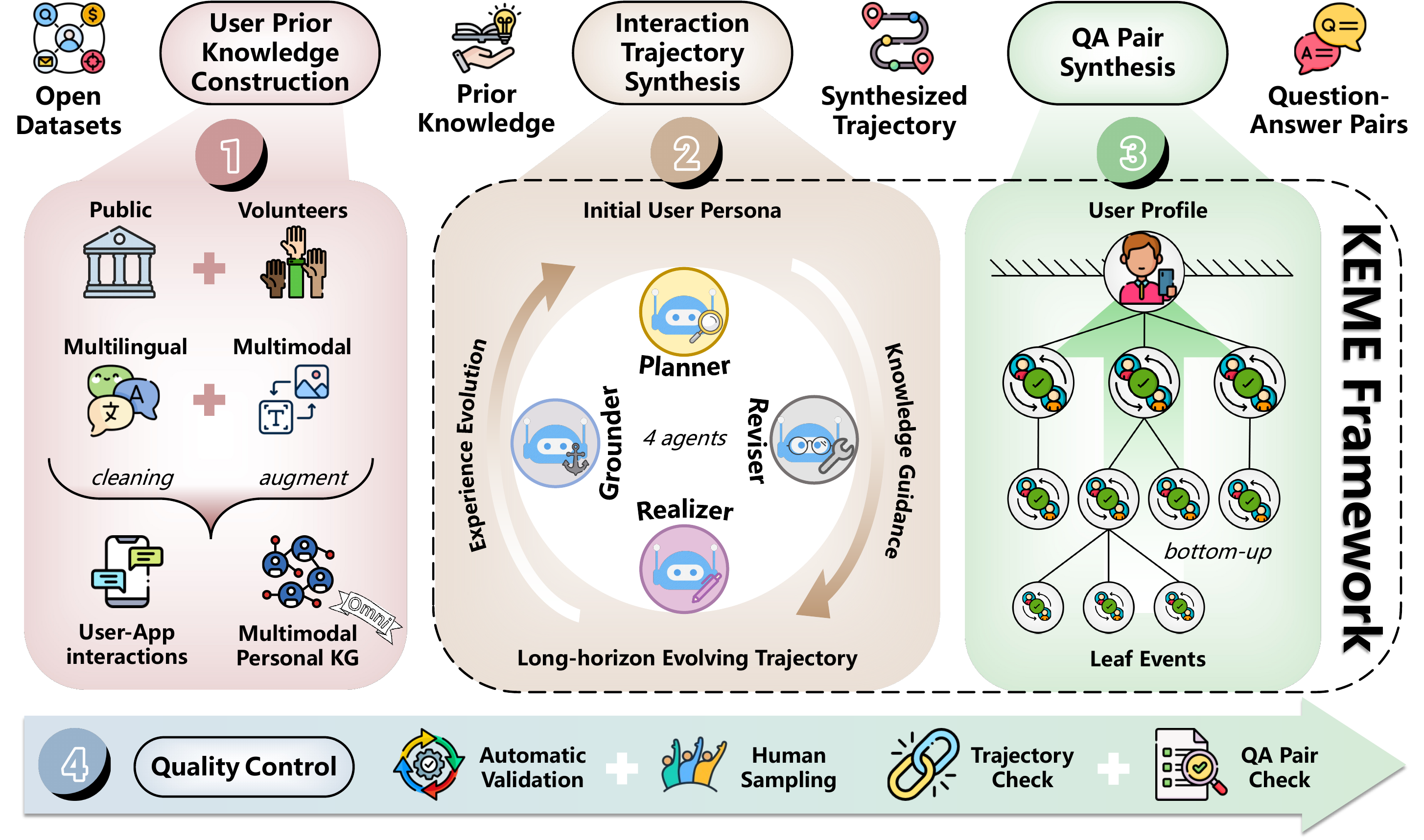}
    \caption{
    Overview of the data synthesis framework. The KEME framework progressively synthesizes long-horizon interaction trajectories and QA pairs from user prior knowledge, with quality control applied throughout the pipeline.
    }
    \label{fig:keme}
    \vspace{-1em}
\end{figure}

In this work, we present a long-horizon data synthesis framework for automatically constructing evaluation data for on-device memory layers. As shown in Figure~\ref{fig:keme}, the framework consists of four major modules: user prior knowledge construction, user interaction trajectory synthesis, question-answer (QA) pair synthesis, and quality control. At the core of the framework is \KEME, a knowledge-guided experience synthesis engine that organizes user priors and knowledge anchors into temporally coherent interaction trajectories and derives QA pairs from them.

\subsection{User Prior Knowledge Construction}

Directly relying on large language models to synthesize user data often leads to limited diversity and insufficient personalization. To address this issue, we introduce user-related prior knowledge as the foundation for data synthesis. Such prior knowledge can be obtained from publicly available user profile datasets such as PersonaHub \citep{Chan2024ScalingSD} and OPeRA \citep{Wang2025OPeRAAD}. 
It can also be collected through interviews or derived from existing open-source datasets such as UltraChat \citep{DBLP:conf/emnlp/DingCXQHL0Z23}. \bench and \omni adopt two different strategies for constructing user prior knowledge. For all data involving real users, we apply appropriate anonymization and desensitization procedures, replacing sensitive information such as names with plausible alternatives.

\paragraph{\bench.} 
\bench defines a dedicated user profile schema and recruits two volunteers with diverse backgrounds. Their user profiles are collected through interviews. In addition, we collect interaction trajectories from the volunteers' OPPO smartphones. 
To strictly comply with application service agreements and privacy standards, our data collection process records only app usage statistics and associated metadata, while explicitly avoiding raw interaction content. 
The collected trajectories also include user-centric contextual signals, such as geographic location, Bluetooth connectivity, and network status. Based on these real but noisy trajectories, we synthesize user-app interactions as supplementary user prior knowledge. Specifically, \bench considers seven types of apps: \textit{Bill}, \textit{Voice Recorder}, \textit{Screen Memo}, \textit{Document}, \textit{Note}, \textit{Calendar}, and \textit{To-Do List}. Each app is associated with a message template. For example, the message template for \textit{Note} is:
\begin{quote}
\texttt{The user creates a note ``[Title]''}\\
\texttt{Content: [Content]}
\end{quote}

As discussed in Section \ref{sec:benchmark_scenarios}, when the user interacts with an app, the app fills the corresponding slots in the message template with the interaction content and forwards the resulting message to the memory layer. Each app is also associated with a structured schema. For each user, we define an app-specific topic distribution based on the user's persona and trajectory. In addition, we assign temporal distributions for \textit{Calendar} and \textit{To-Do List}, and word-count distributions for \textit{Note}, in order to better reflect realistic usage habits. LLMs then generate concrete interaction content through few-shot prompting, taking as input the trajectory, app schema, persona profile, and sampled topic or constraint associated with the message. A notable exception is the synthesis of screen memos, which we construct using the Wikipedia API\footnote{\url{https://wikipedia-api.readthedocs.io/en/latest/API.html\#wikipedia}}. After all messages are synthesized, we group them into user-app interaction sessions according to their timestamps. Specifically, when the time interval between two adjacent messages exceeds two days, the preceding ungrouped messages are segmented into one session. 
This process yields a set of user-app interaction sessions for each user.

\paragraph{\omni.}
\omni also defines a user profile schema. Unlike \bench, this schema additionally includes important events and future plans within a specific year. We recruit eight volunteers and collect their user profiles through interviews. To further increase data diversity, we use the real user profiles as demonstrations and prompt LLMs to generate eight additional virtual user profiles. Moreover, \omni enriches user prior knowledge with a personal relationship graph. For each user, an LLM extracts a user-centered relationship graph from the social relationship information provided in the user profile. Each node in the graph corresponds to a person and is associated with a textual description. Each edge represents a social relationship between two persons. To support visual memory point synthesis, we further generate frontal portrait images for the persons in the graph using a text-to-image model conditioned on their descriptions.

\subsection{\KEME: \underline{K}nowledge-guided \underline{E}xperience synthesis for evolving \underline{ME}mory}

Given user prior knowledge, our goal is to synthesize long-horizon user interaction trajectories that remain temporally coherent and internally consistent. A natural strategy is to decompose this long-horizon synthesis problem into a hierarchy of shorter trajectory synthesis tasks. Besides, some types of prior knowledge can be directly incorporated as part of the final interaction trajectory. For example, the user-app sessions in \bench correspond to interactions that have already occurred and therefore must be preserved. We refer to such prior instances as \textit{knowledge anchors}. 

To this end, we propose \KEME, a knowledge-guided synthesis engine for constructing evolving memory trajectories. \textbf{Guided by user persona knowledge and temporal constraints, \KEME hierarchically organizes knowledge anchors into a unified interaction stream, while progressively synthesizing human-assistant interactions that naturally emerge as the user's experiences unfold over time.} Thus, the trajectory is not statically assembled, but incrementally expanded and refined under accumulated knowledge and past interactions. Beyond trajectory synthesis, \KEME integrates a \textit{bottom-up synthesis algorithm} to produce high-quality QA pairs, enabling systematic evaluation of memory systems under realistic long-horizon scenarios.

\bench and \omni differ in both their sources of user prior knowledge and their observation modalities. As a result, their trajectory synthesis procedures are not identical. \textbf{In the following sections, we present the full \KEME procedure as the core synthesis engine, while directly incorporating the simplified or modality-specific implementations used by \omni where applicable.}

\subsubsection{Formalization and Data Models}
We formalize the synthesis process as a two-stage pipeline. In the first stage, given an initial user persona $\mathcal{P}_{\mathrm{start}}$, a set of knowledge anchors $\mathcal{K}$, and a time horizon $[T_{\mathrm{start}}, T_{\mathrm{end}}]$, \KEME generates a long-horizon evolving trajectory $\mathcal{T}$. In the second stage, the framework constructs a set of QA pairs $\mathcal{Q}$ based on the synthesized trajectory $\mathcal{T}$, a question taxonomy $\mathcal{B}$, and optional auxiliary information such as the final user profile $\mathcal{P}_{\mathrm{end}}$. To ensure robustness, \KEME defines a suite of data models together with corresponding validation logic. Agents in our framework interact with these models through tool calls, which support the creation, modification, and deletion of structured synthesis objects.

The formalization of \KEME transcends the specific scope of \ours series. By defining a generic interface for knowledge anchors $\mathcal{K}$, \KEME provides a compositional synthesis framework that can transform fragmented off-the-shelf data into coherent long-horizon, multi-session interaction trajectories. This capability is particularly important given the scarcity of multi-session data required for training and evaluating agents with long-horizon understanding and cross-session reasoning abilities. We provide further analysis of \KEME in Section~\ref{sec:mobilemem_analysis}. For \omni, the personal knowledge graph is treated as part of the initial persona $\mathcal{P}_{\mathrm{start}}$. Moreover, the synthesis workflow is completely user-predefined. Therefore, the agents do not need to construct trajectories through iterative tool calls over the full set of data models. Instead, each synthesis stage is implemented through direct single-turn prompting with predefined inputs and outputs, while preserving the same high-level principle of prior-grounded, temporally coherent trajectory construction.

\subsubsection{User Trajectory Synthesis}
The trajectory construction in \KEME\ is organized as a \textit{closed-loop process} centered on two principles: \textit{top-down knowledge guidance} and \textit{bottom-up experience evolution}. This design enables long-horizon coherence under anchored knowledge while progressively unfolding a person’s evolving experiences over time. Operationally, \KEME\ instantiates this loop with four specialized agents: a \textit{Knowledge-Guided Planner} $\mathsf{A}_{\mathrm{plan}}$, a \textit{Knowledge Anchor Grounder} $\mathsf{A}_{\mathrm{ground}}$, an \textit{Experience Realizer} $\mathsf{A}_{\mathrm{realize}}$, and an \textit{Experience-Driven Reviser} $\mathsf{A}_{\mathrm{revise}}$.

\paragraph{Top-Down Knowledge Guidance.} 
Trajectory synthesis starts from a root node $r$ that represents the initial persona $\mathcal{P}_{\mathrm{start}}$, together with a time horizon $[T_{\mathrm{start}}, T_{\mathrm{end}}]$ and a set of knowledge anchors $\mathcal{K}^{(r)}$. Given $(\mathcal{P}_{\mathrm{start}}, [T_{\mathrm{start}}, T_{\mathrm{end}}])$, the knowledge-guided planner $\mathsf{A}_{\mathrm{plan}}$ constructs a root-level temporal event graph $\mathcal{G}^{(r)}=(\mathcal{V}^{(r)}, \mathcal{E}^{(r)})$ that partitions the horizon into a small set of coarse-grained life events. Each event is required to be temporally valid\footnote{Its start and end timestamps must lie within the parent node's time horizon.} and compatible with the persona.

The planner then expands the graph in a recursive manner reflecting the hierarchical structure of human experiences. Each event node $n$ can be further expanded into either a finer-grained temporal sub-event graph $\mathcal{G}^{(n)}$  or a leaf-level session $\mathcal{S}^{(n)}$, thereby forming a multi-level decomposition from coarse life phases to concrete interactions. The expansion strategy depends on both the hierarchy depth and the number of grounded sessions associated with the current event. By default, if the current depth reaches $d_{\max}$\footnote{We define the depth of the root node to be 0.}, events are expanded into sessions, whereas at shallower depths they are expanded into sub-event graphs. However, if the number of grounded sessions associated with an event exceeds a predefined threshold, the event is further expanded into a sub-event graph regardless of the current depth. At every recursion level, the constraints inherited from the parent node, which may be semantic, temporal, or otherwise task-specific, together with the current persona state, define the admissible space for descendant synthesis.

Knowledge anchors are integrated through a grounding step whenever the current parent node $u$ carries anchored sessions $\mathcal{K}^{(u)}$. Specifically, the knowledge anchor grounder $\mathsf{A}_{\mathrm{ground}}$ assigns each anchored session in $\mathcal{K}^{(u)}$ to a compatible event node in the corresponding graph $\mathcal{G}^{(u)}$, where compatibility requires that the event time interval fully contains the session interval. When no compatible event exists, the agent revises $\mathcal{G}^{(u)}$ so that every anchor becomes groundable. For each event receiving anchors, the grounded content is summarized into an event-level \textit{compatibility context}. 
This context is propagated to subsequent expansions as a non-contradiction constraint, ensuring anchored knowledge is preserved throughout the hierarchy and preventing synthesized human-assistant interactions from conflicting with the grounded sessions.

\bench and \omni differ slightly in their event schemas and expansion procedures. Each event in \omni records its participants, allowing the pipeline to retrieve their portraits from the personal knowledge graph for image synthesis. It also records the mobile apps involved in the event, facilitating the subsequent synthesis of corresponding mobile app screenshots. \omni uses both hierarchy depth and event duration to determine whether an event should be expanded. Events are classified as short-, medium-, or long-term. Short-term events always remain leaves regardless of their depth. The two benchmarks also represent expansion constraints differently. In \bench, the planner can obtain the information relevant to expansion from the parent event's \texttt{requirement} field. The parent event therefore provides sufficient context for expansion because relevant constraints from earlier events have already been propagated into this field. \omni does not use a \texttt{requirement} field and instead provides the parent event together with all temporally preceding events as context. In addition to these schema and expansion differences, \omni adopts a simplified version of the general \KEME procedure. It constructs only chain-structured temporal event graphs and omits knowledge-anchor grounding.

\paragraph{Bottom-Up Experience Evolution.}
Given a temporal event graph, the knowledge-guided planner $\mathsf{A}_{\mathrm{plan}}$ expands event nodes by following a topological order. Importantly, each expansion produces new concrete outcomes and constraints that are not fully available at planning time. Therefore, after expanding an event, the experience-driven reviser $\mathsf{A}_{\mathrm{revise}}$ updates the remaining unexpanded parts of the corresponding graph. Concretely, $\mathsf{A}_{\mathrm{revise}}$ may add, remove, or adjust future events and dependency edges, to reflect implications of the newly synthesized experience. These refinements aim to resolve emerging inconsistencies, prevent unnatural future transitions, and increase structural richness. This \textit{expansion-refinement alternation} forms a bottom-up feedback loop that continually revises the graph as new experiences are synthesized.

At the leaf level, the experience realizer $\mathsf{A}_{\mathrm{realize}}$ produces concrete sessions. If an event has grounded external sessions, the anchored content is directly adopted as the event outcome, or merged when multiple anchors are assigned, thereby preserving the immutability of anchored knowledge. 
Otherwise, $\mathsf{A}_{\mathrm{realize}}$ synthesizes natural human-assistant dialogues that are consistent with the event context and all inherited constraints. 
Beyond generating interactions, sessions also drive \textit{persona evolution}. 
We represent a persona as a set of dimension-wise attributes with explicit version histories, where each attribute version maintains message-level evidence links. 
Attributes are initially unrevealed in the sense that they have no linked evidence. During session synthesis, only user or system messages are allowed to be linked to persona attributes, turning them into evidence-grounded revealed states. Moreover, if the realized experience implies genuine state changes, such as updated preferences, new goals, or revised habits, the corresponding attributes are updated at the end of the session and recorded with explicit operation logs. Consequently, the synthesized trajectory not only maintains long-horizon coherence under anchored knowledge, but also captures how a person’s experiences incrementally reshape future events and progressively disclose and update the persona over time.

In \omni, session synthesis is conditioned not only on the current event and persona profile, but also on the parent event, when available. Before generating a session, $\mathsf{A}_{\mathrm{realize}}$ first synthesizes a \textit{dialogue goal} and a set of \textit{memory points}. The dialogue goal summarizes the purpose and intended content of the human-assistant interaction, whereas the memory points specify the information that the memory layer is expected to retain. The concrete dialogue is then synthesized according to these intermediate specifications. In particular, \omni distinguishes between textual and visual memory points according to their modalities. A visual memory point remains a textual description from which the corresponding image is subsequently synthesized. Three image synthesis procedures are used depending on the image type. For mobile app screenshots, such as those of music or video applications, the visual memory point provides structured field values that are inserted into predefined HTML templates and rendered into screenshots. For camera photos depicting the user participating in an event with other individuals, the participants specified by the visual memory point are used to retrieve their portraits from the personal knowledge graph. These portraits, together with the memory-point description, condition the photo synthesis process. Other images, such as photos of food or scenery, are synthesized directly from the textual description in the visual memory point. The dialogue itself is synthesized solely from the textual content of the memory points, without taking the generated images as input. Each synthesized message explicitly references the memory points it realizes, enabling the evidence span corresponding to each memory point to be identified within the session. Due to the scale of \omni and the extreme length of its synthesized trajectories, which can exceed two million tokens, the bottom-up feedback loop is omitted to reduce synthesis costs. Nevertheless, \omni retains an implicit form of persona evolution: when expanding an event, the planner conditions on all preceding events within the current temporal event graph. Consequently, the cumulative effects of prior experiences can influence subsequent event synthesis, partially capturing persona evolution without explicitly revising the persona profile through bottom-up feedback.

\subsubsection{Question-Answer Pair Synthesis}
The inputs to question-answer synthesis can be organized as hierarchical tree structures. To fully leverage these structures, \KEME employs a bottom-up algorithm that starts from an initial question type taxonomy $\mathcal{B}_{\mathrm{start}}$ and traverses each tree from its leaves to the root. There are two types of leaf nodes: a profile leaf, corresponding to an individual dimension of the user profile, and a leaf event, corresponding to a concrete event whose expansion yields an interaction session. At each leaf, the agent generates question-answer pairs grounded in the associated profile attributes or session content. An internal node represents either the complete user profile or an event whose expansion yields a temporal event graph. At each internal node, the agent uses the child pairs $\mathcal{Q}_{\mathrm{child}}$ as building blocks to synthesize more complex questions that require reasoning across multiple profile dimensions or interaction sessions. Unused child pairs are randomly sampled and propagated upward together with the newly synthesized complex pairs, allowing higher-level nodes to construct questions over increasingly broader contexts. During synthesis, the agent may also extend the question type taxonomy by adding newly identified question types.

In the \omni synthesis pipeline, the question type taxonomy is predefined and remains fixed throughout synthesis. For each session, question-answer pairs are generated directly from all memory points associated with that session, together with its corresponding event. Because \omni uses a relatively shallow maximum expansion depth, as described in Section~\ref{sec:omni_overview}, it does not apply the bottom-up procedure to synthesize increasingly complex question-answer pairs across hierarchy levels. Moreover, questions that assess changes in the user profile can be synthesized directly from the corresponding memory points. Therefore, \omni does not separately generate question-answer pairs from the complete user profile or its individual dimensions.

\subsection{Quality Control}
We implement a multi-stage quality assurance pipeline to ensure the reliability of the synthesized data. Because the interaction trajectories can be extremely long, no existing automatic metric can reliably assess their global internal consistency. We therefore manually sample trajectory segments and inspect whether their content conflicts with preceding interactions or events. Benefiting from carefully designed prompts and built-in automatic validation mechanisms, this inspection reveals no salient inconsistencies. We observe only a small number of minor numerical discrepancies, such as changes in product prices or the ages of non-user participants. Because these deviations are limited and resemble plausible errors in user-provided information, we retain them as natural noise rather than treating them as structural inconsistencies.

For \omni, quality control additionally covers synthesized images. We first use a multimodal large language model to assess whether each image is consistent with its textual description and filter out images containing clear semantic mismatches, unsafe content, or poor visual quality. For camera photos involving the user or individuals represented in the personal knowledge graph, we additionally use face recognition tools \citep{ArcFace} from the InsightFace library\footnote{\url{https://github.com/deepinsight/insightface}} to compare the synthesized faces against their initial reference portraits. Images whose similarity scores fall below the threshold for same-person identification are rejected. Rejected images are regenerated using either a text-to-image model or an image editing model. To reduce the likelihood of repeated failures, subsequent attempts may use different generation or editing models. We repeat this procedure for a fixed number of attempts and discard images that consistently fail the quality checks. Our manual trajectory inspection also considers the associated images and reveals no salient inconsistencies between the visual content and the surrounding trajectory.

For question-answer pairs, we use an LLM to verify the sufficiency of the supporting evidence and detect questions that are highly similar to previously assessed pairs. Instances identified as unsupported, low-quality, or redundant are discarded. Finally, we manually inspect a random sample of the retained question-answer pairs and find nearly all of them to be correct and adequately supported. We nevertheless observe that some multi-hop questions are less natural, as they appear to combine two independently answerable single-hop questions. Although such questions can be compositionally awkward, their answers remain well-defined and they still evaluate the ability of memory systems to retrieve and integrate evidence across multiple hops. Similar compositional artifacts can also be observed in automatically synthesized datasets such as MultiHop-RAG \citep{Tang2024MultiHopRAGBR}.
\section{MobileMem Dataset}

\subsection{Overview}
For the synthesis of \bench, GPT-5.2 is used as the backbone model for KEME. The maximum expansion depth is set to 2. For each temporal event graph, the minimum and maximum numbers of events are set to 2 and 15, respectively. The maximum number of grounded sessions assigned to an event is set to 1. The maximum number of tokens in the compatibility context for each event is set to 8,000. If the token count exceeds this limit, summarization is applied to reduce the context length. The initial question taxonomy includes single-hop recall, multi-hop reasoning, temporal reasoning, relationship understanding, query-focused summarization, preference-related questions, and unanswerable question recognition. During synthesis, KEME is allowed to further expand this taxonomy with additional question categories. Since preference-related questions and these newly introduced categories occur relatively infrequently in the synthesized data, they are uniformly grouped into the ``Others'' category for analysis. When synthesizing QA pairs, 10 unused child QA pairs are randomly sampled and propagated upward.

\subsection{Experiments}

\subsubsection{Experimental Setup}
\paragraph{Methods.} We evaluate a diverse set of memory systems on \bench, including the long-context memory (Long Context), RAG-style memory systems (NaiveRAG \citep{rag}, HippoRAG2 \citep{HippoRAG2}, and A-MEM \citep{AMEM}), and more complex memory systems, including LangMem \citep{LangMem}, MemOS \citep{MemOS}, EverMemOS \citep{EverMemOS}, Mem0 and its graph-based variant Mem$0^g$ \citep{Mem0}.

\paragraph{Models.} All memory systems use Qwen3-Embedding-4B \citep{Qwen3Embedding} as the embedding model. For Long Context, the context window is set to 128k tokens. For all other memory systems, the number of retrieved memory units is fixed at 30. EverMemOS employs Qwen3-Reranker-4B \citep{Qwen3Embedding} as its reranker. Each memory system is instantiated with two LLM backbones, GPT-4.1-mini \citep{GPT41} and GPT-5.4-mini \citep{GPT54}.

\paragraph{Metrics.} For \bench, we use Qwen3.5-397B-A17B \citep{qwen3.5} as the LLM-as-a-Judge model for automatic evaluation. Additionally, we report the average token overhead incurred by each memory system when it uses LLMs to construct memories for each user trajectory.

\subsubsection{Main Results}

\begin{table*}[t]
\centering
\renewcommand{\arraystretch}{1.00}
\setlength{\tabcolsep}{2.5pt}
\setlength\dashlinedash{1.0pt}
\setlength\dashlinegap{2pt}
\small
\scalebox{0.98}{
\begin{tabular}{@{}lccccccccccc@{}}
\toprule

\multirow{2}{*}{\textbf{Method}}
& \multicolumn{8}{c}{\textbf{Performance}}
& \multicolumn{3}{c}{\textbf{Cost}} \\

\cmidrule(lr){2-9}
\cmidrule(lr){10-12}

& \textbf{Single.}
& \textbf{Multi.}
& \textbf{Temp.}
& \textbf{Rel.}
& \textbf{QFS}
& \textbf{Adv.}
& \textbf{Others}
& \textbf{Overall}
& \textbf{In. Tok.}
& \textbf{Out. Tok.}
& \textbf{Overall}
\\

\midrule

\multicolumn{12}{@{}c@{}}{\textbf{GPT-4.1-mini}}\\

Long Context 
& 56.54 & 51.71 & 53.76 & 58.33 & 47.06 & 51.89 & 60.87 & 54.51
& \textbf{0.00} & \textbf{0.00} & \textbf{0.00} \\

NaiveRAG 
& 38.58 & 31.23 & 29.03 & 29.17 & 20.59 & 60.38 & 47.83 & 37.23
& \textbf{0.00} & \textbf{0.00} & \textbf{0.00} \\

HippoRAG2 
& 85.67 & 79.00 & 61.29 & \textbf{87.50} & 73.53 & 50.94 & 82.61 & 78.85
& 2,307.44 & 495.84 & 2,803.28 \\

LangMem 
& 23.62 & 14.96 & 12.90 & 20.83 & 11.76 & \textbf{76.42} & 39.13 & 24.79
& 3,836.97 & 548.49 & 4,385.46 \\

A-MEM 
& \textbf{86.30} & \textbf{79.79} & \textbf{72.04} & 83.33 & \textbf{76.47} & 44.34 & \textbf{84.78} & \textbf{79.68}
& 4,544.41 & 919.31 & 5,463.72 \\

Mem0 
& 38.11 & 27.30 & 26.88 & 50.00 & 8.82 & 57.55 & 50.00 & 35.63
& 1,472.69 & 597.22 & 2,069.91 \\

Mem$0^g$
& 39.06 & 30.71 & 29.03 & 54.17 & 11.76 & 50.00 & 52.17 & 36.85
& 1,474.46 & 600.36 & 2,074.82 \\

MemOS 
& 72.44 & 61.94 & 52.69 & 75.00 & 50.00 & 50.94 & 76.09 & 65.88
& 3,260.79 & 919.78 & 4,180.57 \\

EverMemOS 
& 68.50 & 57.48 & 55.91 & 83.33 & 55.88 & 47.17 & 76.09 & 62.93
& 7,052.32 & 518.14 & 7,570.46 \\

\midrule

\multicolumn{12}{@{}c@{}}{\textbf{GPT-5.4-mini}}\\

Long Context 
& 49.13 & 42.26 & 35.48 & 45.83 & 50.00 & 34.91 & 54.35 & 45.19
& \textbf{0.00} & \textbf{0.00} & \textbf{0.00} \\

NaiveRAG 
& 40.16 & 32.28 & 26.88 & 45.83 & 23.53 & 42.45 & 58.70 & 37.45
& \textbf{0.00} & \textbf{0.00} & \textbf{0.00} \\

HippoRAG2 
& \textbf{86.14} & \textbf{79.27} & 67.74 & 91.67 & \textbf{85.29} & 50.94 & \textbf{84.78} & \textbf{80.06}
& 2,154.32 & 586.73 & 2,741.05 \\

LangMem 
& 30.71 & 21.26 & 16.13 & 41.67 & 20.59 & \textbf{63.21} & 54.35 & 30.33
& 4,399.06 & 838.12 & 5,237.17 \\

A-MEM 
& 85.67 & 77.17 & \textbf{68.82} & \textbf{95.83} & 73.53 & 43.40 & 82.61 & 78.39
& 9,029.91 & 2,140.53 & 11,170.44 \\

Mem0 
& 52.28 & 31.76 & 31.18 & 45.83 & 14.71 & 35.85 & 56.52 & 42.61
& 1,841.28 & 668.07 & 2,509.34 \\

Mem$0^g$
& 50.71 & 30.97 & 29.03 & 54.17 & 14.71 & 39.62 & 52.17 & 41.77
& 1,848.94 & 696.20 & 2,545.14 \\

MemOS 
& 79.84 & 72.18 & 55.91 & 75.00 & 55.88 & \textbf{63.21} & 82.61 & 74.00
& 4,201.20 & 1,234.40 & 5,435.60 \\

EverMemOS 
& 67.09 & 56.69 & 54.84 & 75.00 & 55.88 & 39.62 & 76.09 & 61.18
& 10,543.65 & 640.15 & 11,183.79 \\

\bottomrule
\end{tabular}}

\caption{\textbf{End-to-end task performance and token cost (in thousands) of memory systems.}
 We evaluate each method with two LLM backbones, GPT-4.1-mini and GPT-5.4-mini. Due to space limitations, the six most frequent question types are displayed, while the remaining question types are grouped into ``Others''. Here, ``Single.'', ``Multi.'', ``Temp.'', ``Rel.'', ``QFS'', and ``Adv.'' denote single-hop, multi-hop, temporal-reasoning, relationship-related, query-focused summarization, and adversarial questions, respectively. ``In. Tok.'' and ``Out. Tok.'' denote the average number of input tokens and output tokens, respectively, used by a memory system to construct memories over one user trajectory. For each backbone, the best result in each column is highlighted in \textbf{bold}.}

\label{tab:main_table_mobilemem}
\end{table*}

\paragraph{Overall performance.}
As shown in Table~\ref{tab:main_table_mobilemem}, A-MEM and HippoRAG2 achieve the best overall performance among all memory systems across the two LLM backbones. One likely reason is that both methods preserve the original conversational information without aggressively compressing, overwriting, or deleting constructed memories. This property is particularly beneficial for \bench, where user trajectories are heterogeneous and many questions depend on fine-grained details in user trajectories. At first glance, this explanation appears to conflict with the relatively poor performance of NaiveRAG, which also preserves the raw conversations. To better understand this gap, we manually compare the memory units retrieved by NaiveRAG, A-MEM, and HippoRAG2. We find that A-MEM and HippoRAG2 achieve substantially higher recall of target memories than NaiveRAG. For A-MEM, the extracted metadata, such as keywords and tags, provides additional signals for distinguishing similar memory units and helps the retriever identify the relevant evidence. HippoRAG2, meanwhile, improves multi-hop memory retrieval by introducing an entity-relation graph and personalized PageRank. These retrieval mechanisms are especially useful in \bench, where the relevant evidence is often surrounded by many distractors. Regarding the impact of LLM backbones, most memory systems obtain comparable or better performance with GPT-5.4-mini than with GPT-4.1-mini. In particular, MemOS and Mem0 show the largest overall gains. In contrast, Long Context performs worse with GPT-5.4-mini. A possible reason is that GPT-5.4-mini has a smaller context window of 400k tokens, compared with the 1 million context window of GPT-4.1-mini. This makes GPT-5.4-mini more vulnerable to performance degradation when it processes long user trajectories.

\paragraph{Task-wise performance.}
We further analyze performance across different question types. Overall, memory systems tend to perform worse as the task requires more evidence. Performance generally drops from single-hop questions to multi-hop questions, and further to query-focused summarization. Query-focused summarization is particularly challenging, as it additionally requires the model to summarize the relevant evidence into a concise answer. Most memory systems also show relatively weak performance on temporal-reasoning questions. This suggests that existing memory construction and retrieval mechanisms still struggle to preserve and recover temporal dependencies among events. For adversarial questions, we observe an interesting pattern. LangMem, despite having the lowest overall performance, achieves the best results on this question type. We find that stronger memory systems often retrieve memories that are only weakly relevant yet highly distracting. This can make the LLM overly confident that an answer exists even when the question is adversarial. Such behavior is consistent with the phenomenon reported by \citet{Lee2026LostIT}.

\paragraph{Token cost.}
We also compare the token cost required by each memory system to construct memories. Long Context and NaiveRAG incur no additional construction cost because they do not invoke an LLM during memory construction. Among the remaining methods, A-MEM and EverMemOS are the most expensive. We further observe that, when using GPT-5.4-mini as the backbone, the construction cost increases for almost all memory systems except HippoRAG2. This increase is especially pronounced for A-MEM, whose total token cost more than doubles. By inspecting the constructed memories, we find that A-MEM with GPT-5.4-mini tends to extract substantially more keywords for each memory unit, which leads to a much higher input and output token cost during memory construction.

\subsubsection{Analysis}
\label{sec:mobilemem_analysis}

\begin{figure}[!h]
    \centering
    \includegraphics[width=\linewidth]{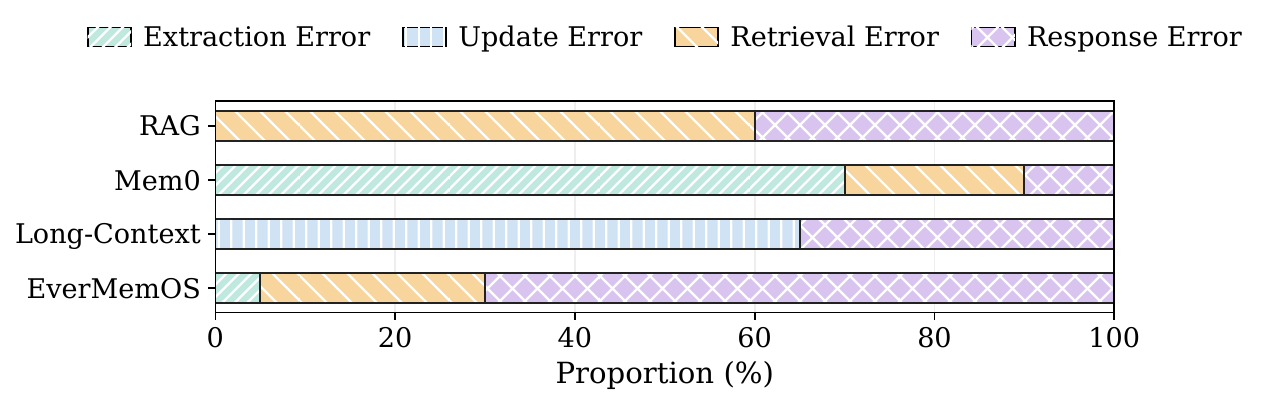}
    \caption{\textbf{Distribution of error types across four memory systems.} We randomly sample 20 failure cases from each system for error analysis.}
    \label{fig:memory_system_error_distribution}
\end{figure}

\paragraph{Memory systems exhibit distinct failure patterns, while answer generation remains a shared bottleneck.} We analyze errors from four memory systems using GPT-4.1-mini as the backbone model. For each system, we randomly sample 20 failure cases. We use deterministic rules to attribute errors in NaiveRAG and Long Context, while applying MemTrace \citep{deng2026memtrace} to Mem0 and EverMemOS. Because automated attribution is not always perfect, we manually inspect the MemTrace results and correct one clear misattribution. Figure~\ref{fig:memory_system_error_distribution} presents the resulting error distributions.

The dominant failure mode differs substantially across systems. For Long Context, 13 of the 20 failures are memory-update errors caused by its limited context window. For NaiveRAG, 12 failures occur because the target evidence is not retrieved. Most Mem0 failures arise during fact extraction. Its extraction prompt primarily focuses on user messages and therefore often misses key facts from app or assistant messages. Other extraction failures result from omitted keywords or fine-grained details. In contrast, most EverMemOS failures occur during answer generation, suggesting that its memory construction and retrieval pipeline is generally effective. MemTrace also reveals two retrieval risks in EverMemOS. First, its sufficiency check may incorrectly trigger another retrieval round. A target document may be retrieved initially but excluded from the top-10 reranked documents used for the sufficiency check. The additional retrieval then introduces distractors into the final reranking stage. Second, the final reranker may directly discard an already retrieved target document. 

Despite their different dominant failure modes, all four systems exhibit answer-generation errors. The downstream language model may be distracted by irrelevant memories, overlook the target evidence, or overgeneralize precise facts and lose important details.

\paragraph{Increasing profile schema granularity improves trajectory diversity, but with diminishing returns under short interaction trajectories.}
To study how profile schema complexity affects trajectory diversity, we define two reduced-dimension profile schemas by down-selecting dimensions from the original schema while preserving broad life-domain coverage. We refer to the original schema as a \textit{fine-grained profile schema} (17 dimensions), and the two reduced variants as \textit{medium-grained} (8 dimensions) and \textit{coarse-grained} (6 dimensions). 
We randomly sample three persona seeds from PersonaHub \citep{Chan2024ScalingSD} and use GPT-4.1 \citep{GPT41} to incrementally expand each seed into the schema. Other schemas are then derived directly from the resulting fine-grained profile.

For trajectory synthesis, we set the maximum number of events to $m_{\max}=10$ to reduce cost, and use an LLM-driven adaptive expansion strategy with a depth cap (event nodes at the maximum depth are forced to be realized as sessions), which typically produces shorter trajectories than non-LLM-driven alternatives. Following prior work \citep{Zhang2025VerbalizedSH}, we measure diversity on user messages using lexical diversity (the distinct bigram ratio computed as the number of unique bigrams divided by the total number of bigrams) and semantic diversity ($1-\hat{s}$, where $\hat{s}$ is the mean pairwise cosine similarity of user message embeddings computed by text-embedding-3-small\footnote{https://developers.openai.com/api/docs/models/text-embedding-3-small}). To mitigate the effect of varying trajectory lengths, for each seed person we compute the minimum number of user messages among the three trajectories and sample that many user messages from each trajectory. 

Table~\ref{tab:additional_analysis:profile_schema_diversity} shows that the medium- and fine-grained schemas yield more diverse trajectories than the coarse-grained schema. However, the fine-grained schema does not provide a commensurate improvement over the reduced schemas. Leveraging \KEME's evidence linking between messages and profile attributes, we observe that after synthesis only about $44\%$ of the fine-grained profile fields are ever mentioned, suggesting that many additional fields remain unused. 
We conjecture that this under-activation is primarily due to the limited trajectory length under our cost-motivated settings, which leaves insufficient opportunity for the generated experiences to surface and utilize the extra fine-grained details.

\begin{table}[h!]
\centering
\small
\setlength{\tabcolsep}{7pt}
\begin{tabular}{lcc}
\toprule
\textbf{Profile Schema} & \textbf{Lexical} $\uparrow$ & \textbf{Semantic} $\uparrow$ \\
\midrule
Coarse-grained  & 0.7238 & 0.4783 \\
\textbf{Medium-grained}  & \textbf{0.7313} & \textbf{0.4932} \\
Fine-grained    & 0.7310 & 0.4875 \\
\bottomrule
\end{tabular}
\caption{\textbf{Trajectory diversity under varying profile schema granularity.} Values are averaged across three seed persons and five length-matched sampling rounds.}
\label{tab:additional_analysis:profile_schema_diversity}
\end{table}

\begin{figure*}[t!]
\centering

\begin{subfigure}[t]{0.8\textwidth}
\centering
\footnotesize
\setlength{\tabcolsep}{7pt}
\begin{tabular}{lcccc}
\toprule
& \multicolumn{2}{c}{NaiveRAG} & \multicolumn{2}{c}{EverMemOS} \\
& Top-$k$=5 & Top-$k$=10 & Top-$k$=5 & Top-$k$=10 \\
\midrule
LongMemEval & 40.00 & 50.00 & 90.00 & 100.00 \\
\textbf{KEME}        & \textbf{20.00} & \textbf{40.00} &  \textbf{80.00} &  \textbf{90.00} \\
\bottomrule
\end{tabular}
\caption{Average performance across two memory systems on LongMemEval and KEME.}
\label{fig:keme_hard_table}
\end{subfigure}

\vspace{0.35em}

\begin{subfigure}[t]{0.57\textwidth}
    \centering
    \includegraphics[width=\textwidth]{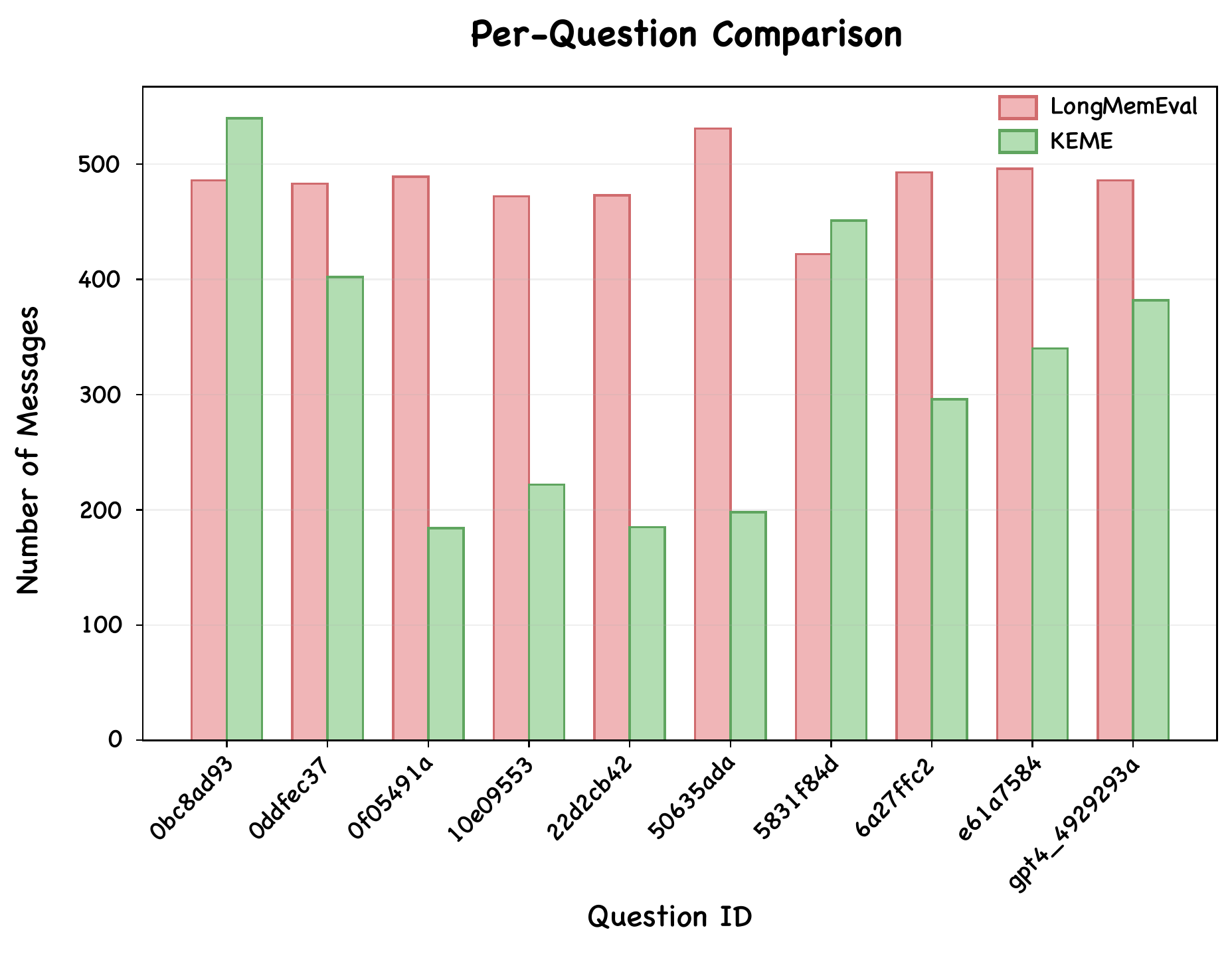}
    \caption{Trajectory length distribution.}
    \label{fig:keme_hard_length}
\end{subfigure}\hfill
\begin{subfigure}[t]{0.39\textwidth}
    \centering
    \includegraphics[width=\textwidth]{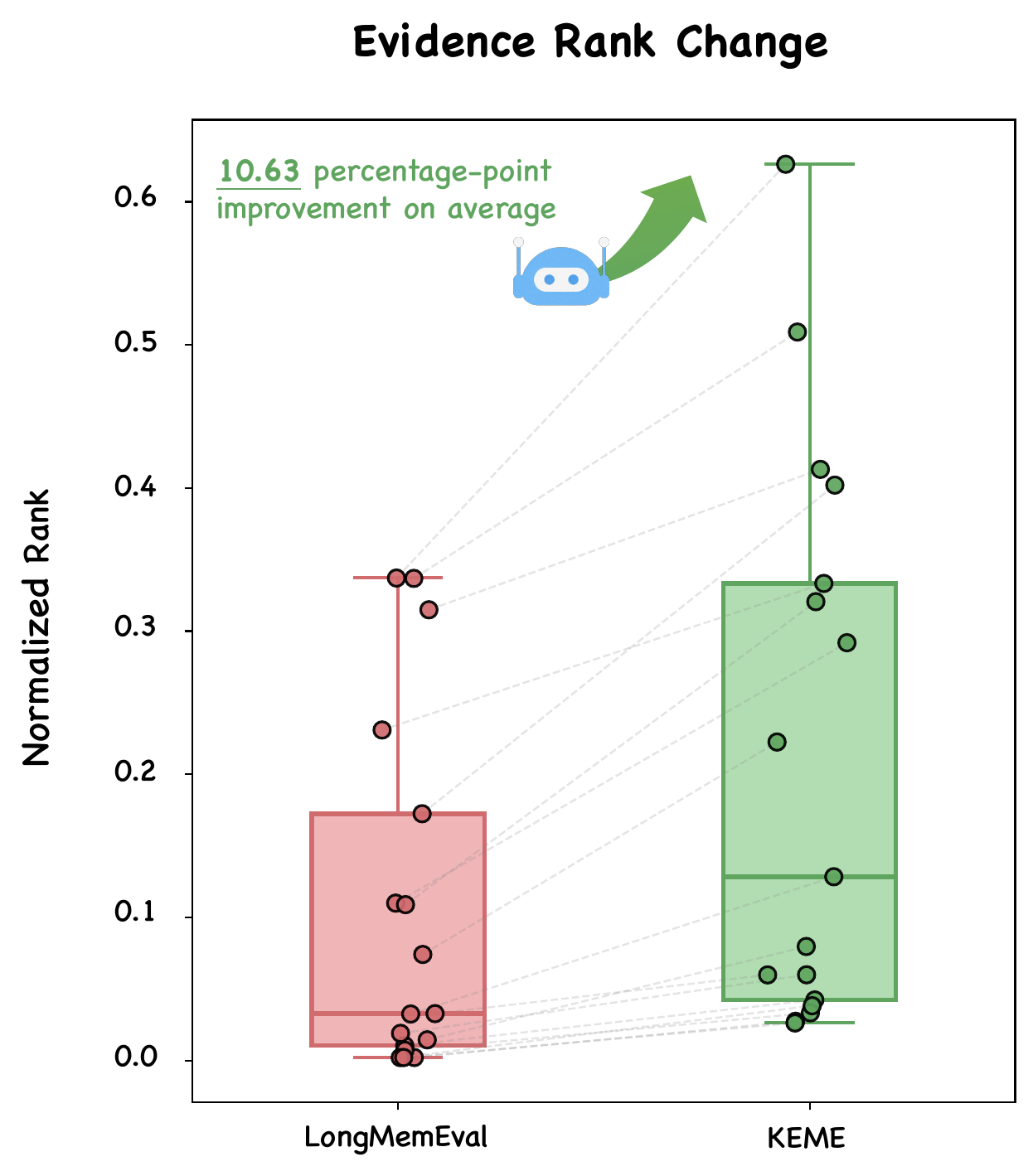}
    \caption{Source evidence rank analysis.}
    \label{fig:keme_hard_rank}
\end{subfigure}

\caption{{\bf Hard-distractor synthesis with \KEME.} 
(a) The top table compares average accuracy (\%) across two memory systems on the original LongMemEval trajectories and the \KEME-synthesized trajectories. When the retrieval budget is set to top-$10$, EverMemOS makes an additional error on the question instance \texttt{50635ada}. When the retrieval budget is reduced to top-$5$, NaiveRAG shows a larger performance drop of $20\%$, failing on \texttt{0ddfec37} and \texttt{gpt4\_4929293a}. (b) The bottom-left panel shows the trajectory length distribution measured by the number of messages. (c) The bottom-right panel visualizes the rank change of source evidence under semantic-similarity-based ranking with NaiveRAG. Each point corresponds to one source evidence and its rank normalized by the total number of memory units. For absolute rank, KEME yields lower ranks for only 4 out of 18 evidence items. However, after normalization, \KEME consistently pushes the answer-supporting evidence toward less favorable retrieval positions by introducing semantically similar distractors, increasing the average normalized rank by 10.63 percentage points. For readability, one evidence (its normalized ranks are approximately 0.9680 for LongMemEval and 1.0000 for \KEME) is omitted from the plot.}
\label{fig:keme_hard_distractor}
\end{figure*}

\paragraph{\KEME can synthesize short but hard trajectories.}
To further show that \KEME can be used beyond benchmark construction, we study whether it can synthesize answer-preserving but retrieval-harder trajectories around existing question-answer pairs. 
We build this setting from LongMemEval where the original benchmark extends trajectories by randomly sampling unrelated extra sessions from existing multi-turn dialogue corpora. Specifically, we first run a weaker baseline, NaiveRAG, and a stronger baseline, EverMemOS, on the original benchmark, and then sample 10 questions, including five cases correctly answered by both systems and five cases correctly answered only by EverMemOS. For each sampled question, we keep the original sessions containing the supporting evidence, and use GPT-5.2 to synthesize a compatible medium-grained persona profile that does not conflict with the given trajectory. We then ask GPT-5.2 to extract the core facts required by the target question and to write guidelines that make the question harder while preserving answer correctness.

Based on these ingredients, we modify only the global system prompt of \KEME. During synthesis, we set $m_{\mathrm{max}}=10$, use an LLM-driven adaptive expansion strategy, and adopt GPT-5.1 \citep{GPT51} as the backbone for all agents. After generation, we manually inspect all 10 synthesized trajectories and verify that the target answers remain unchanged.

Figure~\ref{fig:keme_hard_distractor} summarizes the results. Although the synthesized trajectories are usually shorter than their LongMemEval counterparts, they still make retrieval substantially harder. On the 10 sampled questions, NaiveRAG drops from 40.00\% to 20.00\% at top-$k=5$ and from 50.00\% to 40.00\% at top-$k=10$. EverMemOS also degrades, from 90.00\% to 80.00\% at top-$k=5$ and from 100.00\% to 90.00\% at top-$k=10$. These results indicate that the difficulty introduced by \KEME does not come from simply increasing trajectory length. Instead, \KEME reshapes the surrounding context so that the answer-bearing evidence becomes less favorably positioned for retrieval.

A representative failure occurs for EverMemOS on question \texttt{50635ada}: \emph{``What was my previous frequent flyer status on United Airlines before I got the current status''}. The correct answer is \emph{Premier Silver}. The trajectory contains two key sessions: in September 2022, the user states, \emph{``I actually just hit 20,000 miles... I'm finally eligible for Premier Silver status''}, and in May 2023, the user says, \emph{``I just reached Premier Gold status''}. Manual inspection shows that EverMemOS does preserve the relevant memory in its memory store, but its retrieval stage fails to recover the precise status evidence. Among the final retrieved memories, none explicitly contains either \emph{Premier Silver} or \emph{Premier Gold}. Instead, retrieval is dominated by semantically related but imprecise memories, such as \emph{``comfortably in United's mid-tier elite band''}, as well as unrelated Gold-level mentions from Hilton, Amex, and Marriott. Because these distractors remain highly similar to the query in embedding space, the agentic retrieval process repeatedly returns vague but topically related evidence, preventing the system from identifying the exact status transition and leading to an incorrect answer.

\section{MobileMem-Omni Dataset}

\subsection{Overview}
\label{sec:omni_overview}
\omni is constructed through a multi-stage pipeline powered by GPT-5.1 for persona generation, event construction, dialogue synthesis, and question formulation. The event hierarchy is structured with a maximum depth of two levels: top-level events are first generated, and each is then decomposed into constituent sub-event steps, with the number of steps varying according to event duration and complexity.
For image generation, we employ three types of tools: \textbf{HTML rendering}, \textbf{text-to-image models}, and \textbf{image editing models}. For text-to-image models and image editing models, Seedream is the primary choice. When the generated images fail to pass quality validation, other models \citep{nanopro, gptimage15, qwenimage} are used sequentially as alternatives.
The question set covers seven categories: Single-Hop, Multi-Hop, Knowledge Update, Temporal Reasoning, Implicit Preference, Abstention, and Visual Reasoning, enabling comprehensive assessment of diverse memory capabilities.
The resulting dataset includes 16 user trajectories, 19,060 images, and 7,415 question-answer pairs.


Table~\ref{tab:stats} presents the statistical overview of \omni. The dataset comprises a balanced set of English and Chinese users, each associated with a substantial number of events that span extended temporal horizons. In terms of interaction statistics, the benchmark features long context lengths per user, with multiple sessions and dialogue turns per session. A considerable number of images are distributed across sessions, reflecting the multimodal nature of real-world mobile usage. The question types are distributed across seven reasoning categories, enabling comprehensive assessment of different memory capabilities.

\begin{table}[!htb]
\centering
\renewcommand{\arraystretch}{1.3}
\setlength{\tabcolsep}{0pt}
\small
\begin{tabular*}{0.7\linewidth}{@{\extracolsep{\fill}} l r @{}}
\toprule
\multicolumn{2}{c}{\textbf{User Statistics}} \\
\midrule
Number of Users                 & 16        \\
English Interaction Users       & 8         \\
Chinese Interaction Users       & 8         \\
Total Events                    & 1,589     \\
\midrule
\multicolumn{2}{c}{\textbf{Interaction Statistics}} \\
\midrule
Avg Context Length (tokens/user)& 1.72M     \\
Avg Session Num per user        & 202.6     \\
Avg Dialogue Turns per Session  & 48.2      \\
Total Dialogue Turns            & 155,670   \\
Avg Images per Session          & 5.88      \\
Total Images                    & 19,060    \\
\midrule
\multicolumn{2}{c}{\textbf{Question Type Distribution}} \\
\midrule
Total Questions                 & 7,415     \\
Single-Hop                      & 986       \\
Multi-Hop                       & 1,135     \\
Knowledge Update                & 1,010     \\
Temporal Reasoning              & 773       \\
Implicit Preference             & 1,226     \\
Abstention                      & 1,208     \\
Visual Reasoning                & 1,077     \\
\bottomrule
\end{tabular*}%
\caption{Statistical Overview of \omni.}
\label{tab:stats}
\end{table}



\begin{table*}[!t]
\centering
\renewcommand{\arraystretch}{1.00}
\setlength{\tabcolsep}{2.5pt}
\small
\begin{tabular*}{\textwidth}{@{\extracolsep{\fill}}lccccccccc@{}}
\toprule
& \multicolumn{7}{c}{\textbf{Task}} & \multicolumn{2}{c}{\textbf{Overall}} \\
\textbf{Method} & \textbf{Single.} & \textbf{Multi.} & \textbf{Know.} & \textbf{Temp.} & \textbf{Abst.} & \textbf{Pref.} & \textbf{Vis.} & \textbf{Judge} & \textbf{F1} \\
\midrule

\multicolumn{10}{c}{\textbf{Textual Memory Methods}} \\
\midrule
\multicolumn{10}{c}{\textbf{GPT-5.4-mini}} \\
Long Context & 17.24 & 10.22 & 21.48 & 19.28 & 18.13 & 34.01 & 12.16 & 19.14 & 20.50 \\
NaiveRAG & 24.85 & 20.44 & 16.53 & 19.02 & 7.37 & 8.56 & 14.5 & 15.4 & 10.4 \\
LangMem & 21.81 & 19.21 & 27.52 & 29.88 & 12.00 & 49.02 & 14.95 & 24.94 & 7.46 \\
Mem0 & 26.88 & 25.90 & 27.33 & 27.43 & 7.12 & 43.96 & 15.04 & 24.73 & 9.63 \\
LightMem & \underline{37.42} & \underline{32.95} & \underline{38.22} & \underline{38.94} & 19.21 & 46.25 & \underline{25.81} & \underline{33.8} & \textbf{21.5} \\
EverMemOS & \textbf{46.96} & \textbf{40.44} & \textbf{46.83} & \textbf{43.47} & 7.04 & \textbf{60.28} & \textbf{34.08} & \textbf{39.41} & 12.16 \\
M\textsuperscript{2}A (w/ Caption) & 27.59 & 29.60 & 22.67 & 16.95 & 4.06 & 32.14 & 19.50 & 21.86 & \underline{21.05} \\
\midrule

\multicolumn{10}{c}{\textbf{Qwen3-VL-8B-Instruct}} \\
Long Context & 4.06 & 4.32 & 5.05 & 9.19 & 6.29 & 6.53 & 6.78 & 5.93 & 12.81 \\
NaiveRAG & 5.07 & 5.64 & 2.77 & 5.30 & 7.45 & 0.73 & 2.41 & 4.15 & 4.38 \\
LangMem & 17.44 & 14.19 & 18.71 & 17.08 & 13.82 & 30.18 & 8.17 & 17.25 & 2.43 \\
Mem0 & 25.89 & 23.23 & 31.81 & 23.70 & 13.01 & \underline{50.49} & 16.79 & 26.71 & 8.77 \\
LightMem & 30.12 & 26.43 & 29.70 & 23.16 & \underline{20.78} & 36.05 & 22.19 & 27.08 & 16.88 \\
EverMemOS & 28.09 & 27.22 & 27.82 & 26.26 & 13.82 & 31.97 & 19.22 & 24.76 & 3.55 \\
M\textsuperscript{2}A (w/ Caption) & 25.96 & 26.70 & 21.88 & 17.85 & 5.46 & 20.80 & 17.21 & 19.21 & 17.53 \\
\midrule

\multicolumn{10}{c}{\textbf{Multimodal Memory Methods}} \\
\midrule
\multicolumn{10}{c}{\textbf{GPT-5.4-mini}} \\
Multimodal Long Context & 34.9 & 20.4 & 30.9 & 31.3 & 14.0 & 17.2 & 21.9 & 24.4 & 18.3 \\
SigLIP + NaiveRAG & 23.1 & 16.83 & 19.70 & 18.1 & 4.8 & 33.8 & 9.6 & 18.00 & 10.0 \\
UniversalRAG & 22.6 & 17.6 & 22.1 & 21.2 & 4.9 & 39.0 & 7.6 & 19.3 & 9.5 \\
M\textsuperscript{2}A  & 1.52 & 1.23 & 2.87 & 6.47 & \textbf{24.44} & 16.39 & 3.63 & 8.68 & 12.12 \\
\midrule
\multicolumn{10}{c}{\textbf{Qwen3-VL-8B-Instruct}} \\
Multimodal Long Context & 16.7 & 11.6 & 18.3 & 19.0 & 1.4 & 9.3 & 13.5 & 12.8 & 12.2 \\
SigLIP + NaiveRAG & 10.9 & 9.9 & 9.8 & 11.9 & 15.56 & 17.5 & 4.4 & 11.6 & 4.6 \\
UniversalRAG & 14.8 & 12.3 & 15.9 & 11.5 & 13.1 & 26.6 & 3.2 & 14.2 & 4.6 \\
M\textsuperscript{2}A & 0.82 & 1.06 & 2.33 & 8.98 & 16.42 & 4.91 & 2.75 & 5.40 & 10.54 \\
\bottomrule
\end{tabular*}
\caption{\textbf{Question-answering performance of methods on \omni.} Scores for task columns are evaluated using LLM-as-a-Judge. Here, ``Single.'', ``Multi.'', ``Know.'', ``Temp.'', ``Abst.'', ``Pref.'', and ``Vis.'' denote single-hop, multi-hop, knowledge-update, temporal-reasoning, abstention, implicit-preference, and visual-reasoning questions, respectively. ``Judge'' denotes the overall LLM-Judge score, and F1 measures word overlap with the ground-truth answer.}
\label{tab:main_table}
\end{table*}

\subsection{Experiments}

\subsubsection{Experimental Setup}
\label{sec:exp_setting}

\paragraph{Methods.} We evaluate a range of memory methods on \omni, covering both textual and multimodal memory methods.
For textual memory methods, images are converted into text using GPT-5.1 to generate high-quality captions. These methods include Long Context, NaiveRAG \citep{rag}, as well as memory frameworks including LangMem \citep{LangMem}, Mem0 \citep{Mem0}, LightMem \citep{LightMem}, and EverMemOS \citep{EverMemOS}.
For multimodal memory methods, we evaluate approaches that directly process visual inputs\footnote{For Multimodal Long Context, images are resized to $256 \times 256$; for others, images are resized to $512 \times 512$.}. These include Multimodal Long Context as a baseline, as well as several multimodal RAG approaches, such as multi-channel RAG, which integrates NaiveRAG for text retrieval and SigLIP2 \citep{SigLIP} for image retrieval, along with UniversalRAG \citep{UniversalRAG}. We also include the multimodal memory framework M\textsuperscript{2}A \citep{M2A} with both caption-based and image-based inputs.

\paragraph{Models.} The all-MiniLM-L6-v2 model \citep{MiniLM} is used as the default text embedding model. For all retrieval-based methods, the default retrieval size is set to $K=15$. Each approach is implemented with two MLLM backbones, GPT-5.4-mini and Qwen3-VL-8B-Instruct.

\paragraph{Metrics.} We adopt two metrics to evaluate answer quality. LLM-Judge uses Qwen3-14B to assess correctness, taking the question, reference answer, model prediction, and relevant evidence as input and outputting a binary judgment. We select Qwen3-14B as the judge model, which achieves an average agreement rate of 80.7\% with human evaluation on a sampled set of 1,320 examples. F1 measures word overlap between the generated answer and the ground-truth answer.

\subsubsection{Main Results}
\paragraph{Overall performance.} As shown in Table~\ref{tab:main_table}, we observe a performance hierarchy across different families of approaches. Specialized memory-augmented frameworks consistently outperform long context and RAG-type methods across almost all tasks and backbone models.
When examining the same method applied to different backbone models, we also observe an advantage of GPT-5.4-mini over Qwen3-VL-8B-Instruct across almost all methods. For example, EverMemOS with GPT-5.4-mini achieves an LLM-Judge score of 39.41\%, substantially higher than the 24.76\% obtained with Qwen3-VL-8B-Instruct. M\textsuperscript{2}A with textual captions also substantially outperforms its multimodal variant using raw images.

\paragraph{Task-wise performance.}
On both GPT-5.4-mini and Qwen3-VL-8B-Instruct, multimodal long context with images substantially outperforms long context with image captions on visual reasoning tasks, achieving 21.9\% compared to 12.16\% and 13.5\% compared to 6.78\%, respectively.
Multi-hop, abstention, and visual reasoning tasks are the most challenging categories.
For abstention tasks, most methods score below 20\%, indicating that memory systems struggle to recognize when they lack sufficient information, consistent with prior work \citep{DBLP:journals/corr/abs-2510-09033, DBLP:conf/icml/ChengSLZYLLH0Q24}.
Furthermore, visual reasoning and multi-hop tasks also prove highly challenging, suggesting that current memory methods lack both fine-grained visual-semantic understanding and the ability to reason across multiple pieces of information.

\begin{figure}[htbp]
    \centering
    \begin{minipage}{0.45\textwidth}
        \centering
        \includegraphics[width=\linewidth]{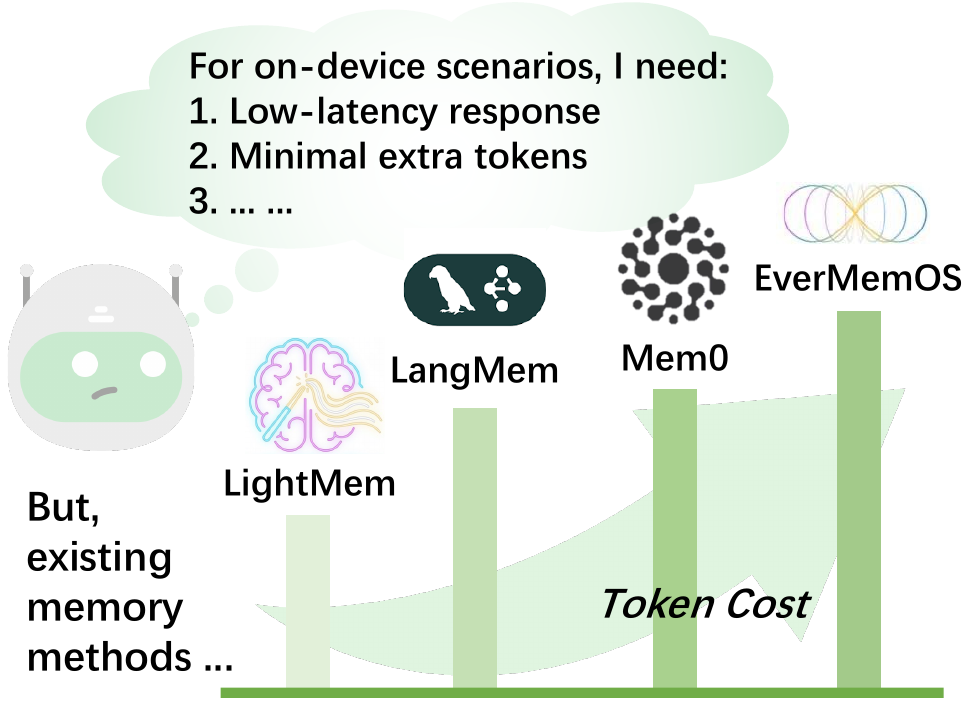}
        \caption{Existing memory methods are slow and resource-intensive, hence unsuitable for on-device scenarios.}
        \label{fig:token_cost}
    \end{minipage}
    \hfill
    \begin{minipage}{0.45\textwidth}
        \centering
        \includegraphics[width=\linewidth]{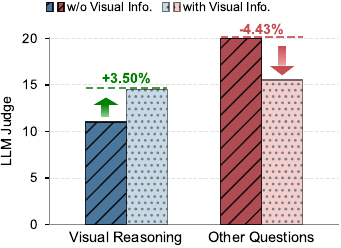}
        \caption{LLM-Judge changes in NaiveRAG: captions improve visual reasoning but degrade other text-oriented questions.}
        \label{fig:visual_ablation}
    \end{minipage}
\end{figure}

\paragraph{Token cost.}
As shown in Figure~\ref{fig:token_cost},
most current memory methods require substantial token costs. EverMemOS achieves the best performance, yet its token cost remains relatively high. LightMem maintains reasonable performance while reducing token consumption. Future memory methods should be more lightweight and token-efficient for on-device deployment.

\subsubsection{Analysis}

\begin{figure*}[!ht]
\centering
\includegraphics[width=0.99\textwidth]{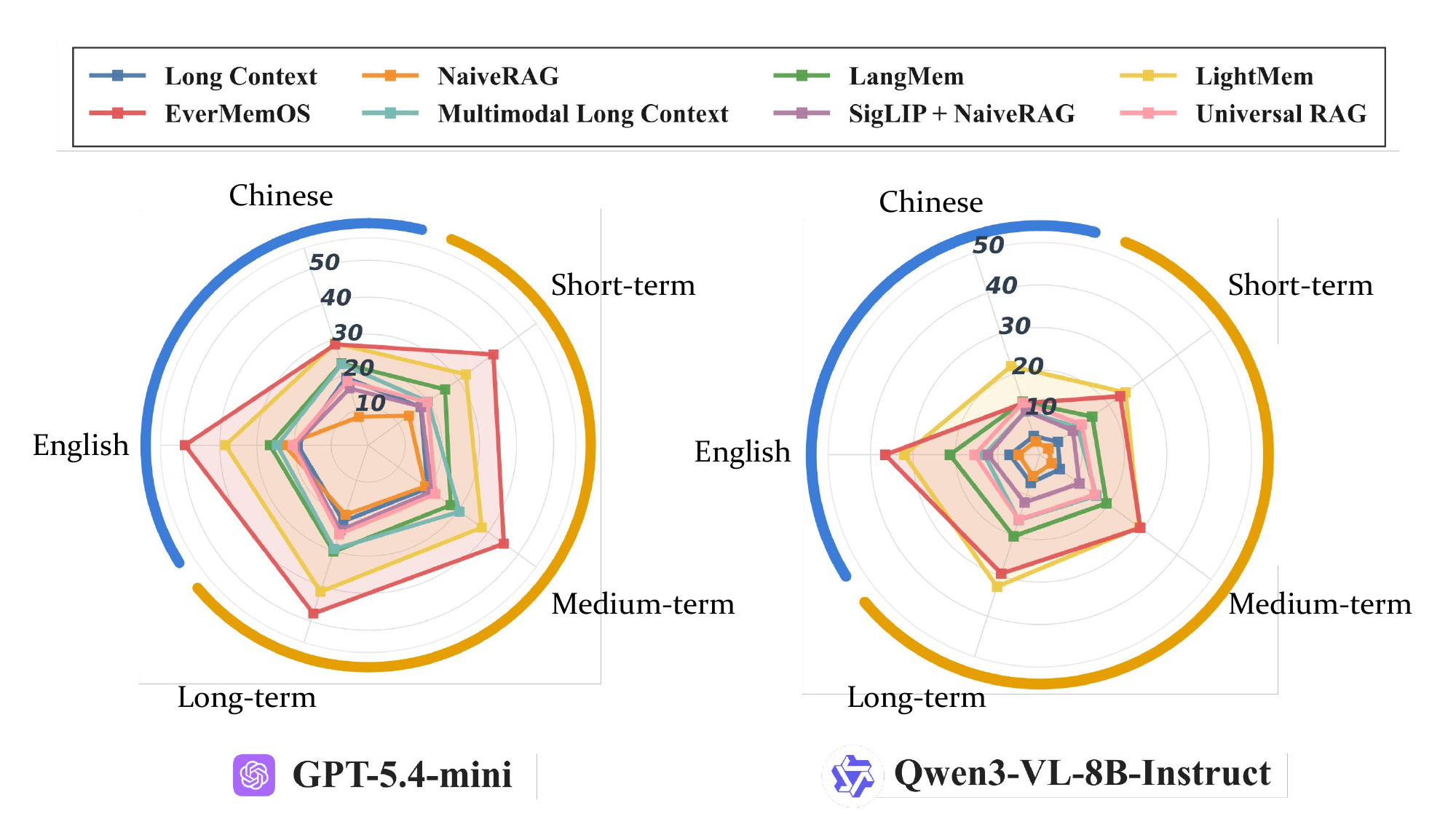}
\caption{Radar chart of event type LLM-Judge performance (\%). Events are divided by interaction language: Chinese, English; and by event span: short-term, medium-term, long-term.}
\label{fig:event_type}
\end{figure*}

\paragraph{Performance varies with event time span.}
As indicated by the three time-span metrics in Figure~\ref{fig:event_type}, memory method performance improves as the event span increases. Short-term events are the most difficult, while longer events yield better results. This may be because longer events provide richer contextual information that aids retrieval and reasoning, whereas shorter events offer limited context and require more precise memory matching, resulting in relatively poorer performance.

\paragraph{Performance gaps persist across languages.}
As indicated by the metrics in Figure~\ref{fig:event_type}, we observe a notable performance gap between English and Chinese across all memory methods. Specifically, most methods achieve lower scores on Chinese questions compared to their English counterparts. This disparity likely stems from the fact that memory mechanisms and methods are primarily designed and optimized for English, resulting in weaker memory capabilities, semantic alignment, and retrieval quality for other languages. This finding highlights the need for more balanced multilingual support in memory methods.

\paragraph{Captions do not guarantee performance gains.}
As shown in Figure~\ref{fig:visual_ablation}, incorporating image captions into NaiveRAG yields a trade-off: performance improves on visual-reasoning questions but declines on text-oriented questions. This indicates that for visual-reasoning tasks, captions provide rich contextual information that aids memory retrieval. However, for text-oriented questions, the inclusion of visual information introduces redundancy and interference that can distract from purely textual reasoning. Therefore, effective memory systems require more fine-grained visual representation modeling rather than relying solely on coarse image captions.

\section{Applications}

\begin{figure*}[!ht]
\centering
\includegraphics[width=0.98\textwidth]{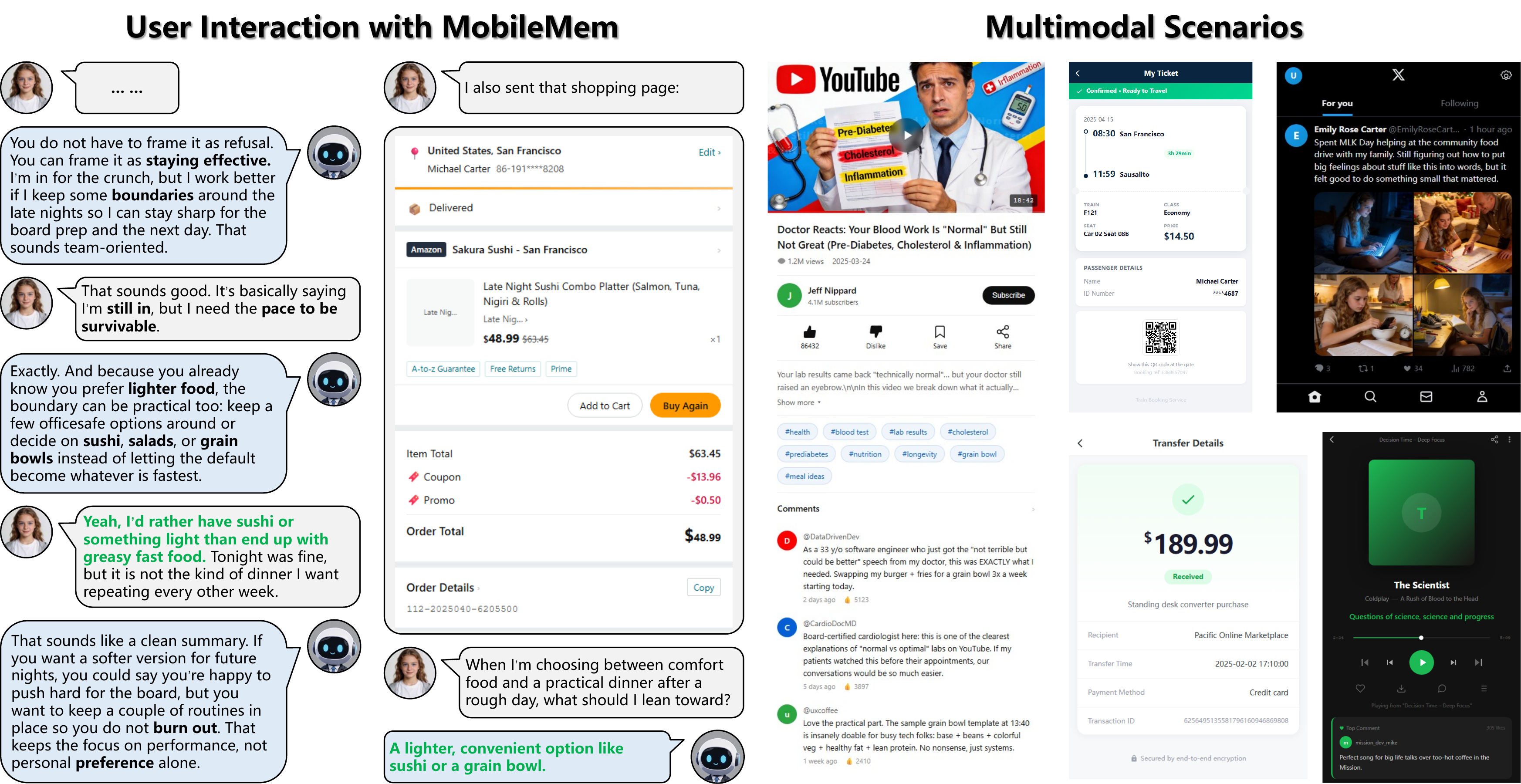}
\caption{Illustrative examples from \omni, covering user interactions and multimodal scenarios with diverse image types.}
\label{fig:case}
\end{figure*}

\paragraph{Running Example in MobileMem.}
Figure~\ref{fig:case} presents a panoramic view of \omni with conversation-level examples spanning user interactions and multimodal scenarios, illustrating the breadth and diversity of the dataset at fine-grained dialogue resolution. We highlight the practical value of \ours from multiple angles, including concrete application scenarios, standardization, user memory value, ecosystem integration, and on-device intelligence.

\paragraph{Application Scenarios.}
\ours is built upon OPPO's native on-device memory ecosystem and covers diverse sources of long-term memory, including Calendar, Photos, Notes, Documents, To-Do List, Voice Recorder, Breeno Memory, and Video Memo. These heterogeneous memory sources collectively capture the temporal information, visual content, textual records, browsing behaviors, multimedia consumption, and personal activity traces generated throughout a user's long-term use of a mobile device, thereby providing a realistic and continuous user memory foundation for the next generation of mobile AI with long-term memory capabilities.

\begin{remarkbox}
\textbf{Health Management and Medication Reminder.}
An elderly user routinely records blood pressure and blood glucose measurements and has kept past medical examination reports and doctor's advice. Over the past few months, after each follow-up visit, the user has noted the next appointment time in Calendar, jotted down the doctor's medication adjustments in Notes, also searched for contraindications of a common medication via the browser, and saved photos of lab reports and prescriptions in Photos. One early morning, suddenly feeling slightly dizzy, the user opens the phone and asks the assistant, \textit{"What medication was I allergic to before? Please check whether any of the medicines I'm currently taking contain it."} By matching the allergy history in the health records with the current medication list, the assistant immediately provides a clear risk alert.
\end{remarkbox}
This scenario demonstrates the value of long-term personal memory in health management and reflects \ours's evaluation capability for cross-source memory association, multimodal information fusion, long-term factual memory, temporal reasoning, and complex memory retrieval.

\begin{remarkbox}
\textbf{Itinerary Planning for a Foreign Trip.}
A tourist planning a trip to Chengdu, China, has browsed numerous travelogues and guides in advance, casually bookmarked several articles, and has become increasingly drawn to the mountains and waters along the Western Sichuan Ring Route. The night before departure, the user says to the assistant, \textit{"Based on everything I've bookmarked, what's a good two-day route around Chengdu?"} Instead of generically recommending Kuanzhai Alley and Jinli, the assistant integrates the content of the bookmarked guides with the user's previously searched transportation modes and accommodation preferences, and produces a concrete two-day itinerary: departing from downtown Chengdu on the first day, passing through Dujiangyan to Yingxiu, and on the second day crossing Balang Mountain Pass to view Mount Siguniang before returning in the evening.
\end{remarkbox}
This scenario illustrates that long-term memory can support not only information retrieval but also cross-source knowledge integration, user preference modeling, multi-hop reasoning, and long-term personalized planning. \ours provides a realistic, continuous, and reproducible evaluation environment for such capabilities.

\begin{remarkbox}
\textbf{Personal Watchlist and Drama Tracking.}
A habitual drama watcher always writes detailed reviews in Notes, accompanied by several memorable screenshots, and has accumulated an archive of impressions covering hundreds of films and TV shows over a long period. Having just finished a new drama today, the user feels it resonates with a previously recorded work but cannot recall the specifics. The user asks the assistant, \textit{"Based on my past drama tracking records, organize everything related to this one and write a long comparative review."} Without resorting to online search, the assistant carefully sifts through the fragmented impressions accumulated over the years in Notes, identifies works similar in style or theme, systematically compares them from several angles item by item, and generates a review with opinions, source material, and traces of personal memory.
\end{remarkbox}
This scenario highlights the importance of long-term memory for interest modeling, cross-temporal knowledge association, and personalized generation, and also reflects the evaluation value of \ours in long-term knowledge organization, multimodal retrieval, and generative reasoning.

\begin{remarkbox}
\textbf{Work Review and Mid-Year Summary.}
A working professional is preparing a mid-year summary. Over the past six months there have been countless meetings, repeatedly revised proposals, and substantial reading of technical articles and industry news, yet it feels like the learning has been diffuse and hard to organize. The user says to the assistant, \textit{"Help me review my work from the past six months. Based on the last tech sharing session, sort out what I've mainly been catching up on and where I've been stuck at work."} The assistant integrates the timelines of all meetings and project milestones from Calendar, the revision histories of reports and proposals from Documents, the fragmented reflections in Notes, and the themes of technical articles browsed through screen memory, distills a clear learning trajectory, contrasts work tasks with actual output, pinpoints the bottlenecks where the user repeatedly got stuck, and generates a well-structured mid-year review summary.
\end{remarkbox}
This scenario illustrates the application potential of long-term memory in knowledge management, continuous learning, and personal growth analysis, and showcases the systematic evaluation value of \ours for core capabilities such as long-term temporal reasoning, cross-source information fusion, complex memory organization, and personalized summary generation.

\paragraph{Standardization.}
A key characteristic of \ours is that it is constructed from authentic smartphone memory rather than synthetic interaction logs or manually curated user profiles. The benchmark unifies foundational memory, cognitive memory, preference memory, temporal reasoning, and visual reasoning within a single evaluation framework, while grounding assessment in approximately one year of real user activities. \ours is built upon heterogeneous memory sources generated by native mobile applications, including calendars, albums, notes, documents, to-do lists, bills, voice memos, screen memories, video memories, and other user-generated records. These complementary sources naturally cover structured records, semi-structured documents, free-form text, images, and multimodal content, providing substantially richer memory distributions than previous long-term memory benchmarks. Furthermore, \ours establishes standardized data schemas, annotation protocols, question taxonomies, evaluation metrics, and benchmark version management, providing a common foundation for future research on personalized AI memory and serving as a practical reference for industrial benchmarking.

\paragraph{User Memory Value.}
At its core, \ours treats long-term multimodal user memory as the fundamental unit of personalization. Unlike conventional question answering benchmarks that primarily evaluate isolated factual recall, \ours evaluates whether AI assistants can continuously accumulate, retrieve, integrate, and reason over evolving user memories across extended temporal horizons. By modeling authentic user trajectories, multilingual interactions, dynamic preferences, and real-world events, the benchmark encourages the development of AI assistants capable of delivering context-aware, proactive, and continuously adaptive assistance. Such an evaluation paradigm reflects practical mobile assistants that interact with users over months rather than individual conversations, shifting personalized AI from generic response generation toward persistent user understanding while preserving user privacy.

\paragraph{Ecosystem.}
\ours is designed as a unified benchmark for evaluating long-term memory in mobile AI assistants rather than a standalone memory system. Its modular architecture seamlessly integrates with the existing mobile agent ecosystem, including task understanding, memory retrieval, question answering, web search, response safety, and execution modules. Such a design enables researchers and practitioners to independently optimize memory retrieval algorithms, retrieval-augmented generation (RAG) pipelines, memory middleware, long-context reasoning, and agent execution while maintaining a unified evaluation protocol. Consequently, \ours supports reproducible comparison across different memory architectures and facilitates the integration of long-term memory evaluation into real-world development pipelines. The benchmark further complements continuous learning frameworks by providing standardized evaluation throughout the lifecycle of memory-enhanced mobile assistants, thereby accelerating iterative development from research prototypes to production systems.

\paragraph{On-Device Intelligence.}
\ours shifts long-term memory evaluation from qualitative capability demonstrations to measurable, comparable, and continuously improvable assessment. The benchmark provides comprehensive metrics for retrieval accuracy, temporal consistency, preference understanding, multimodal reasoning, and long-horizon knowledge integration, enabling objective comparison among different memory architectures and retrieval strategies. These quantitative evaluations provide valuable guidance for optimizing on-device memory middleware under realistic deployment constraints, including limited computation, storage capacity, latency requirements, and privacy preservation. As foundation models increasingly migrate from cloud services to edge devices, \ours offers an effective framework for evaluating reliable, efficient, and privacy-aware memory systems that continuously evolve through long-term user interactions.

\paragraph{Future Outlook.}
The current version of the on-device memory benchmark focuses on long-term horizons and naturally supports expansion toward a broader ecosystem of personalized AI. Future versions will incorporate memory sources from wearables, tablets, personal computers, smart home devices, and in-vehicle systems, enabling a unified evaluation of ecosystem-wide lifelong memory. We also plan to extend multilingual coverage, increase the diversity of user populations, and introduce more challenging reasoning tasks involving long-term planning, collaborative memory, continual personalization, and cross-modal knowledge integration. We expect the mobile memory benchmark to evolve into a comprehensive benchmark that bridges academic research and industrial deployment, while fostering the development of trustworthy, personalized, and continually evolving AI assistants.

\section{Conclusion and Future Work}
We introduce \ours, a comprehensive benchmarking framework for evaluating on-device memory systems in realistic mobile environments. Our knowledge-grounded synthesis pipeline enables privacy-preserving generation of long-horizon, temporally coherent interaction trajectories, upon which we construct a large-scale benchmark covering heterogeneous interaction types and diverse reasoning categories.

Future directions include extending \ours to additional languages and modalities, such as audio, video, and sensor data, to better reflect the richness of real-world mobile usage; establishing online evaluation protocols for continuous assessment of evolving memory systems; further strengthening privacy preservation through advanced synthesis techniques; and investigating the distinct bottlenecks observed in memory construction, retrieval, updating, and answer generation, and developing targeted solutions accordingly. We will maintain \ours as an open platform, welcoming community contributions to advance on-device memory intelligence.

\paragraph{Limitations.}
First, the data may inevitably contain noise, inaccuracies, or inconsistencies, particularly when long-horizon experiences are reconstructed from incomplete or imperfect prior knowledge. 
Although we introduce quality-control mechanisms, fully eliminating such errors remains challenging. 
Second, our current user modeling is still relatively coarse and may not capture the full complexity of users' evolving preferences, habits, goals, and contextual factors. 
A more comprehensive and dynamically evolving user model could further improve the fidelity and diversity of the synthesized experiences. 
Third, the memory systems and LLMs evaluated in this work do not yet fully represent the rapidly evolving capabilities of state-of-the-art personal agents. 
Consequently, our current evaluation may not capture all aspects of long-term memory performance in future agentic systems.


\section*{Contributions}
We sincerely thank all the units, organizations, and individuals who contributed to this report.

\textbf{Partner Organizations} are Guangdong OPPO Mobile Telecommunications Corp., Ltd. and the OpenKG community.

\textbf{Core Authors (equal contribution)} are Yida Xue, Xinle Deng, and Xiangyuan Ru. 

\textbf{Data and Application} are Yijun Chen, Buqiang Xu, Mingjun Mao, Xinjie Liu, Chi Zhang, Ruobin Zhong, Jizhan Fang, Haoming Xu, Chen Jiang, Haibo Li, Qiqi Chen, Nandi Zhang, and Ming Chen.

\textbf{Legal and Compliance} are Fan Jia, Da Lu, Zilong Ye, Xiaohua Wen, Yunxia Lei, Zhentong Li, and Yuhao Zhai.

\textbf{Technical Leads} are Yuchen Eleanor Jiang, Lizhong Wang, Jason Wang, Li Zeng, Xiaofan Zhang, Minxin Tu, Yongrui Chen, Zequn Sun, Tianxing Wu, Meng Wang, Xiang Chen, Zhen Bi, Wei Hu, Haofen Wang, Guilin Qi, Huajun Chen, and Ningyu Zhang.

We hope this technical report provides a valuable reference for researchers and practitioners working on mobile AI agents and on-device memory systems.
\bibliography{references}


\end{document}